# Advancing machine fault diagnosis: A detailed examination of convolutional neural networks

Govind Vashishtha [1], Sumika Chauhan[1*], Mert Sehri[2], Justyna Hebda-Sobkowicz[1], Radoslaw Zimroz[1], Patrick Dumond[2] and Rajesh Kumar[3]

[1]Faculty of Geoengineering, Mining and Geology, Wroclaw University of Science and Technology, Na Grobli 15, 50-421 Wroclaw, Poland
[2]Department of Mechanical Engineering, University of Ottawa, 161 Louis Pasteur, Ottawa, Ontario, Canada
[3]Precision Metrology Laboratory, Department of Mechanical Engineering, Sant Longowal Institute of Engineering and Technology, Longowal 148 106, India
* Corresponding author, Email address: sumi.chauhan2@gmail.com

**Abstract:** The growing complexity of machinery and the increasing demand for operational efficiency and safety have driven the development of advanced fault diagnosis techniques. Among these, convolutional neural networks (CNNs) have emerged as a powerful tool, offering robust and accurate fault detection and classification capabilities. This comprehensive review delves into the application of CNNs in machine fault diagnosis, covering its theoretical foundation, architectural variations, and practical implementations. The strengths and limitations of CNNs are analyzed in this domain, discussing their effectiveness in handling various fault types, data complexities, and operational environments. Furthermore, we explore the evolving landscape of CNN-based fault diagnosis, examining recent advancements in data augmentation, transfer learning, and hybrid architectures. Finally, we highlight future research directions and potential challenges to further enhance the application of CNNs for reliable and proactive machine fault diagnosis.

## 1. Introduction

The relentless pursuit of operational efficiency, safety, and economic viability in modern industries has propelled the development of sophisticated machinery capable of performing complex tasks. However, this increased complexity comes at a price: the potential for malfunctions and failures grows exponentially, impacting production, safety, and overall system performance [1][2]. Timely and accurate fault diagnosis becomes crucial for mitigating these risks, enabling proactive maintenance, and minimizing downtime and associated costs [3][4].

Traditional fault diagnosis methods often rely on a combination of expert knowledge, rule-based systems, and signal-processing techniques. While these approaches have proven effective in specific scenarios, they face limitations in handling complex fault patterns, noisy data, variable working conditions, and evolving machine characteristics [5][6]. The rapid advancements in artificial intelligence (AI) and machine learning (ML) have opened new avenues for fault diagnosis, offering data-driven solutions capable of learning complex relationships and making accurate predictions.

Machine fault diagnosis has witnessed a steady evolution of approaches, driven by the ingenuity of researchers and engineers. These methods can be broadly categorized into four main groups: physical model-based, signal processing-based, machine learning-based, and hybrid approaches [7][8]. Physical model-based methods rely on a deep understanding of the machine's inner workings. However, creating accurate models for modern complex machinery, particularly in dynamic and noisy environments, presents a significant challenge. Furthermore, these models often lack flexibility, proving difficult to update with real-time monitoring data [9][10]. Signal processing-based methods focus on enhancing fault characteristic information through advanced noise reduction and filtering techniques. This approach requires a strong understanding of the relevant equipment and its frequency characteristics. Additionally, establishing a robust theoretical foundation and mathematical basis for fault representations is crucial for its success. Machine learning-based methods, representing data-driven approaches, have gained immense momentum in the modern industrial landscape. While classical machine learning models, such as support vector machines (SVMs) and k-nearest neighbors (kNNs), have achieved impressive results, they face limitations when confronted with the demands of modern industry [11][12]. For instance, the need for manual feature extraction and selection hinders their effectiveness in analyzing complex big data, while still requiring domain expertise. Their shallow structures struggle to effectively mine high-dimensional features. Separate feature mining and decision-making processes lead to inefficient optimization, impacting performance. As sensor diversity and machine complexity increase, traditional algorithms struggle to deliver satisfactory results in the face of growing data dimensionality and dynamics. Hybrid approaches aim to leverage the strengths of different methods, combining the knowledge-driven aspects of physical models with the data-driven power of machine learning and the analytical prowess of signal processing. This combined approach holds immense promise for tackling the complex challenges of machine fault diagnosis in modern industrial settings.

Deep learning, a rapidly evolving branch of machine learning, has experienced remarkable growth and success across diverse fields like image recognition and language processing. This surge in popularity is driven by a confluence of factors [13]. For instance, Deep learning's inherent ability to process massive datasets and learn complex features, coupled with innovative architectural advancements, has fueled its rapid adoption [14][15]. The explosion of industrial big data, breakthroughs in hardware and IoT technology, and the increasing demand for intelligent solutions in various domains have further propelled the advancement of deep learning [16]. Naturally, this wave of innovation has also impacted machine fault diagnosis. Over the past five years, deep learning has revolutionized fault diagnosis, leading to the development of various models, including deep autoencoders, deep belief networks (DBNs), recurrent neural networks (RNNs), and CNNs.

Among these various ML techniques, CNNs have emerged as a powerful tool for machine fault diagnosis. CNNs, renowned for their prowess in image recognition and other visual tasks, exhibit a unique ability to extract spatial features from complex data, making them particularly well-suited for analyzing signals generated by machines. This ability to decipher the intricate patterns present in vibration signals, acoustic emissions, and sensor readings empowers CNNs to identify and classify faults with remarkable accuracy. The general framework of convolutional network-based fault diagnosis (CNFD) is shown in Fig. 1.

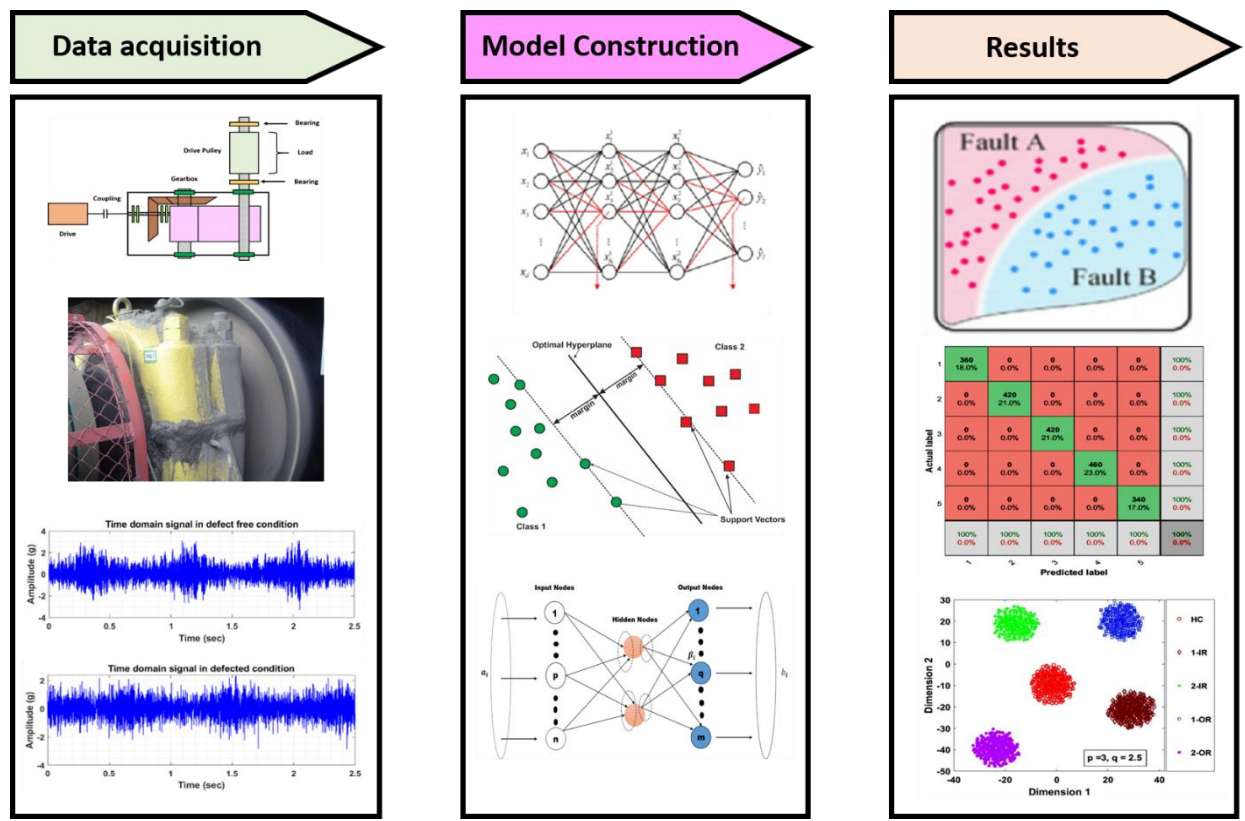

**Fig. 1** General framework of CNFD

This comprehensive review delves into the application of CNNs in machine fault diagnosis, providing a comprehensive overview of the theoretical foundations, architectural variations, and practical implementations that have propelled this field forward. The aim is to shed light on the strengths and limitations of CNNs in this domain, discussing their effectiveness in handling various fault types, data complexities, and operational environments. Furthermore, the evolving landscape of CNN-based fault diagnosis is explored, examining recent advancements in data augmentation, transfer learning, and hybrid architectures. Finally, the review highlights future research directions and potential challenges to further enhance the application of CNNs for reliable and proactive machine fault diagnosis.

## 2. Data acquisition

The success of training convolutional neural networks for fault diagnosis hinges on the availability of high-quality data. This data acquisition process involves two crucial steps: sensor selection and layout, followed by data sampling and storage.

Advances in sensor technology have led to the use of various sensors for monitoring mechanical conditions, including accelerometers, current sensors, and built-in encoders. These sensors capture comprehensive monitoring information for fault diagnosis [17]. Vibration analysis has emerged as the most widely used monitoring technique, rapidly evolving over recent years. While effective, it faces limitations in practical application such as vibration data often suffering from interference due to transmission paths and environmental noise, resulting in a low signal-to-noise ratio. Moreover, vibration data is not sensitive to low-frequency signals, making them unsuitable for monitoring low-speed machinery [18]. Installation of vibration sensors can be challenging in high-temperature, high-pressure, or enclosed environments. To address these limitations, alternative sensors, such as infrared imaging (non-contact measurement) and built-in encoders (improved signal-to-noise ratio and low-frequency response), offer viable solutions. Selecting appropriate sensors requires a holistic approach, considering factors like equipment type, working environment, monitoring objects, and operating conditions. Optimizing sensor placement is crucial for capturing a maximum amount of health information and minimizing the impact of transmission paths and interference.

Data collection requirements include considerations for both hardware and software. For hardware parameters, the natural frequencies of machine components need to be calculated to identify machinery health state classes. Once all machinery components are identified, then a selection for accelerometer sampling frequencies can be made by data scientists. Additionally, when collecting data for ML fault diagnosis it is essential to understand frequency harmonics. Frequency harmonics are multiples of the machine component's natural frequency. Although there is currently no standard for machine component data collection in ML diagnosis, the identification of machine component frequencies will allow for easier data acquisition and sensor selection. Therefore, it is recommended that data scientists and ML researchers work together to identify the number of machine component natural frequency harmonics required to reach a threshold diagnosis accuracy, which in turn, will allow a reduction in the number of samples required, the sampling frequency, and will reduce computer storage requirements. Additionally, software parameters to consider include hyper-parameter optimization such as batch size, input signal length, overlap, stride, learning rate, and number of epochs. Each parameter can be optimized to increase diagnosis accuracy to determine fault classifications. Diagnosis accuracy can also be increased with the creation of synthetic data to complement existing available datasets when there are limited amounts of data. However, the creation of synthetic data requires knowledge of machinery component natural frequencies to imitate realistic datasets as closely as possible. These data collection practices can hopefully bridge the gap between research and industry, allowing ML algorithms to diagnose faults with less computational hardware requirements, in turn tempting internet of things (IoT) companies to integrate ML into viable condition monitoring solutions.

Data sampling is carried out using a data acquisition system. The acquired data is stored on hard drives or cloud platforms for subsequent analysis and utilization [19]. Thus, it can be suggested that obtaining high-quality data for training CNNs involves careful sensor selection, strategic sensor placement, and robust data sampling and storage processes. This comprehensive approach ensures that the network receives an optimal input, facilitating accurate and reliable fault diagnosis. While the process of collecting data for machine fault diagnosis appears straightforward, acquiring high-quality data in real industrial settings presents significant challenges:

- **Limited fault data:** Obtaining fault data is often difficult because machines are typically not allowed to operate under fault conditions for safety and operational reasons [20][21].

- **Time-consuming life-cycle data:** Gathering data that spans the entire life cycle of a machine, from a healthy state to failure, is a time-consuming, expensive, and often impractical endeavor due to the extended operational lifespan of many machines [22][23].

To address these limitations, some institutions have generously made their lab created datasets publicly available for research and application. This review introduces several such public datasets, providing researchers and engineers with valuable resources for evaluating the performance of their fault diagnosis approaches.

## 2.1. Case Western Reserve University (CWRU) bearing fault dataset

The Case Western Reserve University (CWRU) bearing fault dataset is a widely recognized and valuable resource in machine fault diagnosis. This dataset, meticulously collected and curated at CWRU, provides a rich collection of vibration signals from various bearing conditions, including healthy bearings and bearings with different types and severities of faults [24]. It encompasses a range of operating conditions, such as different motor speeds and applied loads, making it a comprehensive and realistic representation of real-world scenarios. The CWRU dataset, as shown in Fig. 2, has been instrumental in advancing research in fault diagnosis, particularly in the development and validation of data-driven models, such as those based on CNNs [25]. Researchers can use this dataset to train and test their algorithms, compare different approaches, and benchmark their performance. However, it must be acknowledged that this dataset was not created explicitly for training and testing machine learning algorithms and if not considered carefully, can often lead to resulting algorithm performance that does not reflect a meaningful outcome [26].

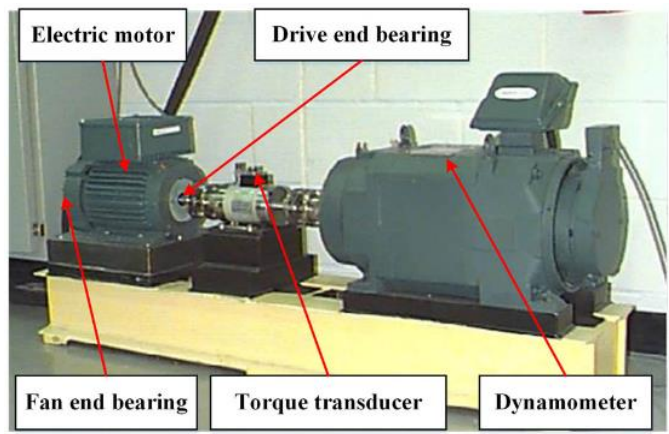

**Fig. 2** Test rig of bearing used in CWRU [26]

## 2.2. University of Ottawa (UORED-VAFCLS) bearing fault dataset

To facilitate machine fault detection, the CWRU dataset can be integrated with the University of Ottawa Rolling Element Dataset – Vibration and Acoustic Faults under Constant Load and Speed conditions (UORED-VAFCLS), as shown in Fig. 3. This dataset includes vibration and acoustic data. Conditions include healthy, as well as ball, cage, inner race, and outer race bearing faults [27]. To replicate realistic conditions, loads were applied to a healthy bearing until failure occured in an accelerated test format. The dataset is helpful for transfer learning since it includes five bearings of each fault type from two distinct bearing manufacturers. The UORED-VAFCLS is particularly well suited as a complement to the CWRU benchmark.

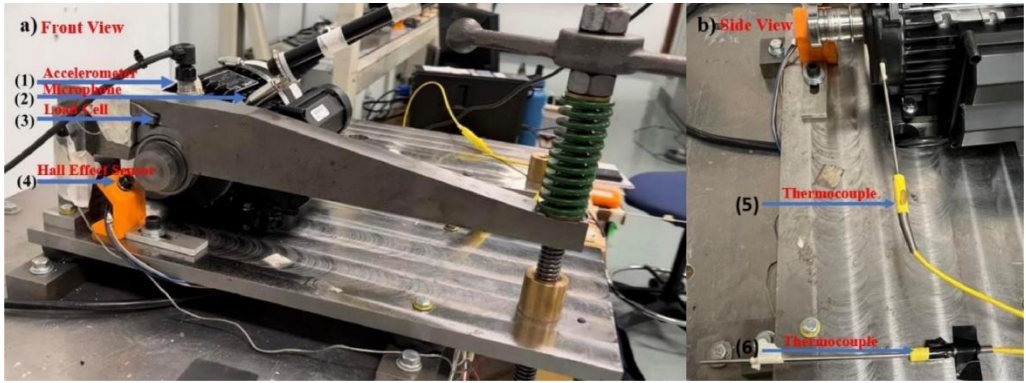
Fig. 3 UORED-VAFCLS bearing test rig [27]

### 2.3. Huazhong University of Science and Technology (HUST) bearing fault dataset

The Hanoi University of Science and Technology (HUST) dataset contains bearings with different sizes and shapes collected on the test rig shown in Fig. 4. These different bearings include the 6204, 6205, 6206, 6207, and 6208 [28]. The dataset also includes inner race faults, outer race faults, ball faults, and combination faults. The dataset also allows for transfer learning applications due to the different dimensions of the bearings included. More promising is the possibility of further improving algorithm results via data fusion of the HUST dataset with the data from the CWRU and UORED-VAFCLS datasets. In the development of algorithms, researchers in machine learning can develop their algorithms using these concatenated datasets in their training and testing processes, which supplement other benchmarks already available in the literature.

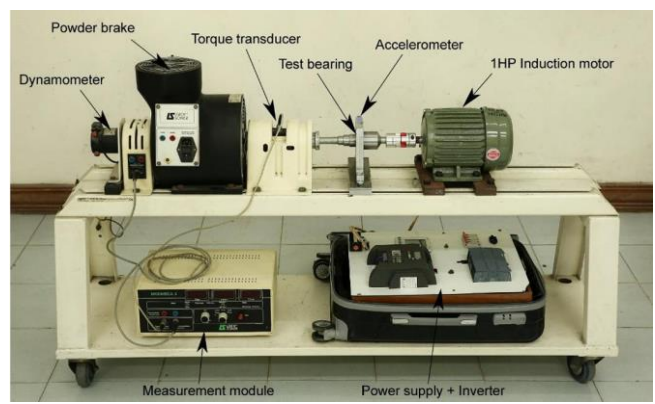
Fig. 4 HUST bearing test rig [28]

### 2.4. University of Ottawa (UOEMD-VAFCVS) induction motor fault dataset

The University of Ottawa Electric Motor Dataset – Vibration and Acoustic Faults under Constant and Variable Speed Conditions (UOEMD-VAFCVS) test rig is shown in Fig. 5. This dataset, also collected at the University of Ottawa, includes vibration and acoustic data. Machinery fault classes include healthy, rotor unbalance, rotor misalignment, stator winding faults, voltage unbalance, bowed rotor, broken rotor bars, and faulty bearings [29]. This dataset contains artificially seeded faults created by Spectra Quest. The dataset is helpful for induction motor fault diagnosis since it includes different motor speeds (constant and variable) and load conditions for each fault type. Additionally, machine learning researchers can easily combine this dataset with the UOEMD-VAFCVS dataset during training and testing of algorithms, which can increase deep learning algorithm fault detection for induction motors.

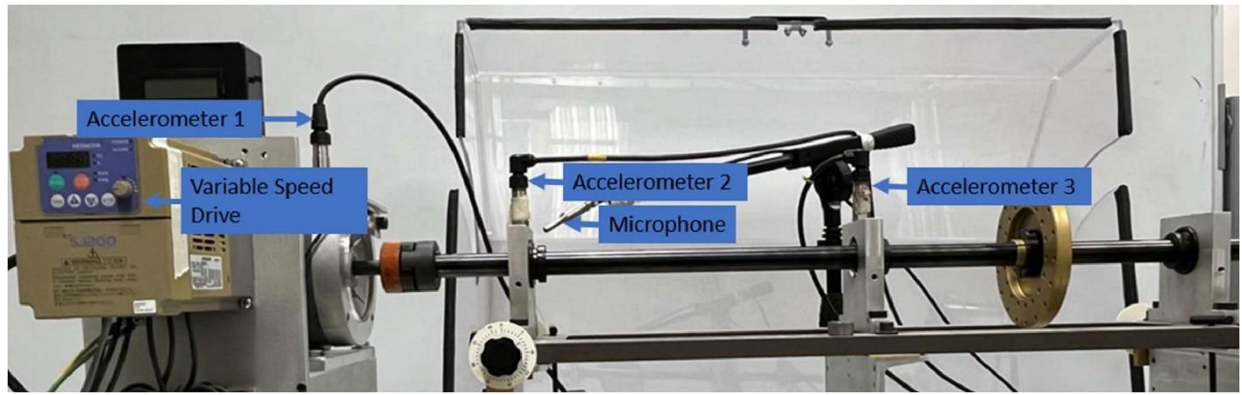

Fig. 5 UOEMD-VAFCVS induction motor test rig [29]

### 2.5. PHM 09 gearbox fault dataset

The PHM 09 gearbox fault dataset, as shown in Fig. 6, is a valuable resource in prognostics and health management (PHM), providing a comprehensive collection of data for gearbox fault diagnosis and prediction. Collected during a PHM competition in 2009, the dataset features vibration signals from a gearbox subjected to various operating conditions and fault scenarios, including healthy operation, tooth chipping, bearing damage, and misalignment [30][31]. This dataset offers a realistic representation of gearbox behaviour under various conditions, allowing researchers to develop and evaluate models for fault detection, classification, and prognosis. The diversity of fault types and operating conditions in the PHM 09 dataset makes it an excellent benchmark for evaluating the robustness and accuracy of different PHM algorithms.

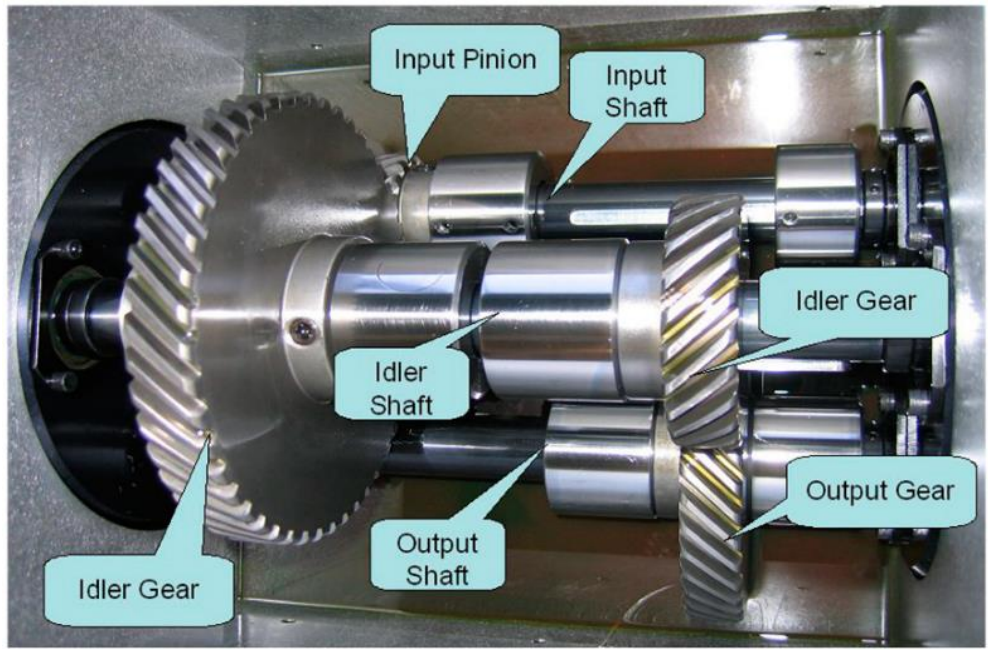

Fig. 6 PHM 09 gearbox test rig [32]

## 2.6. Paderborn dataset

The Paderborn dataset is a valuable resource for research in machinery fault diagnosis, particularly for analyzing acoustic emission (AE) signals. Collected at the University of Paderborn, as shown in Fig. 7, this dataset contains AE signals from a variety of gearboxes operating under different load and speed conditions [33]. It features various natural and seeded fault scenarios, including tooth damage, bearing defects, and misalignment, providing a rich and realistic representation of gearbox behavior. The Paderborn dataset is notable for its high sampling rate and the inclusion of multiple sensor channels, allowing researchers to investigate the complex relationships between AE signals and specific fault types [34]. Its availability has facilitated the development and validation of advanced AE-based fault diagnosis techniques. The dataset's comprehensive nature has made it a valuable benchmark for evaluating the effectiveness of different fault diagnosis approaches.

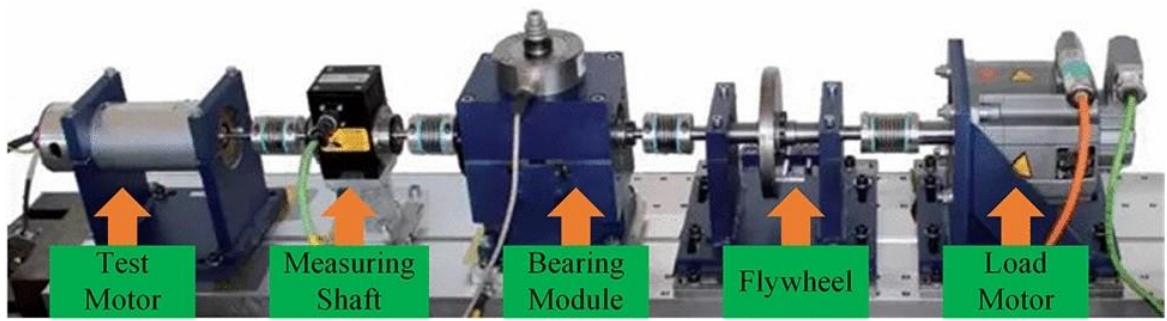

Fig. 7 Paderborn test rig [35]

## 2.7. IMS bearing dataset

The IMS bearing dataset, as shown in Fig. 8, collected from the Inner Motor Shaft (IMS) bearings in automotive engines, serves as a crucial resource for researchers in the field of machine fault diagnosis. This dataset captures the intricate vibration signals emanating from these bearings, often subjected to demanding operating conditions [36]. It encompasses various fault scenarios, such as wear, pitting, and cracks, offering a comprehensive representation of potential bearing failures. The IMS bearing dataset is particularly valuable due to its focus on a specific type of bearing commonly used in automotive applications, allowing researchers to develop and evaluate models tailored for this critical component [37]. The dataset's realistic representation of real-world conditions makes it an ideal platform for testing and validating advanced diagnostic techniques to analyze vibration signals and predict impending failures.

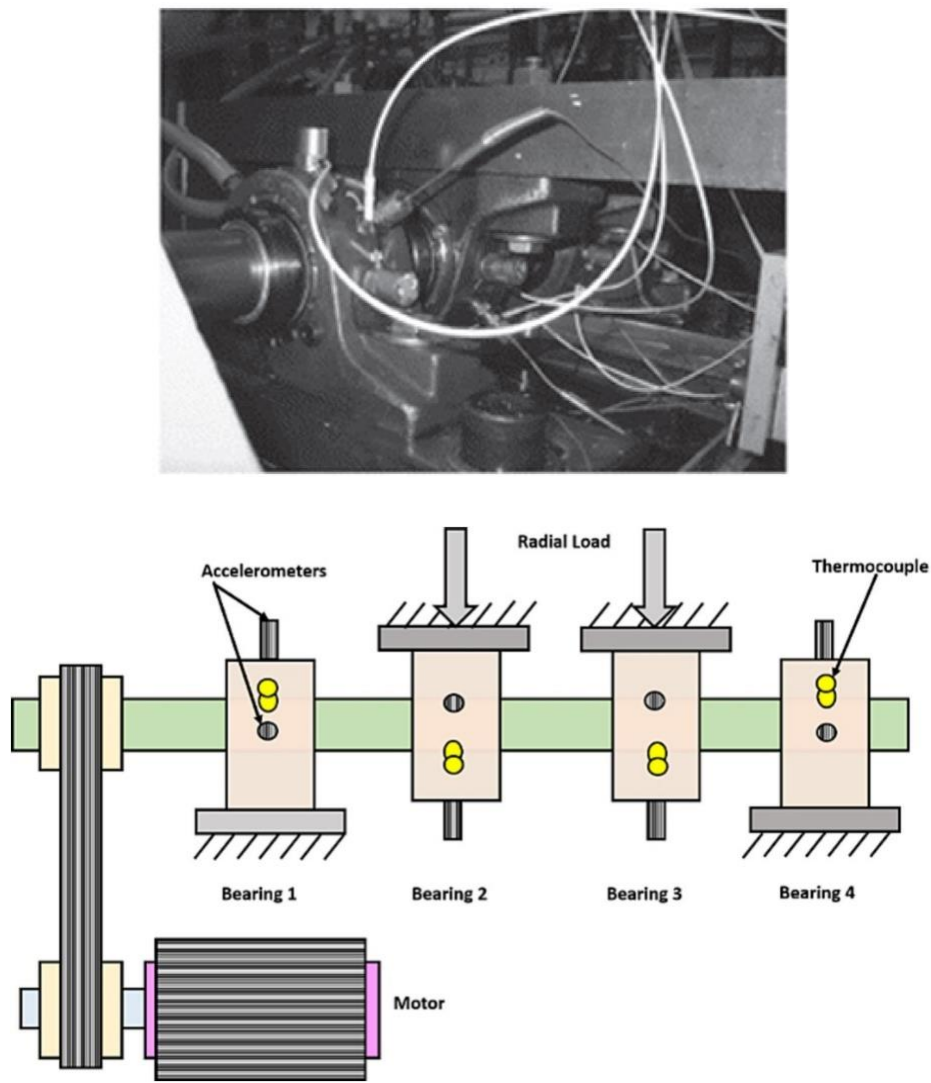

**Fig. 8** IMS bearing experimental setup [38]

## 2.8. C-MAPSS dataset

The C-MAPSS dataset, as shown in Fig. 9, is a valuable resource in the field of PHM, providing a comprehensive collection of data for turbofan engine health monitoring. This dataset, developed by NASA, features sensor readings from a simulated turbofan engine under various operating conditions and degradation scenarios [39]. It includes detailed information about engine parameters, such as fan speed, core speed, and engine pressure ratios, allowing researchers to develop and evaluate models for predicting remaining useful life (RUL) and detecting potential failures. The C-MAPSS dataset offers a realistic representation of turbofan engine operation and degradation, making it an essential benchmark for evaluating the performance of various PHM algorithms, especially those employing deep learning techniques like CNNs [40].

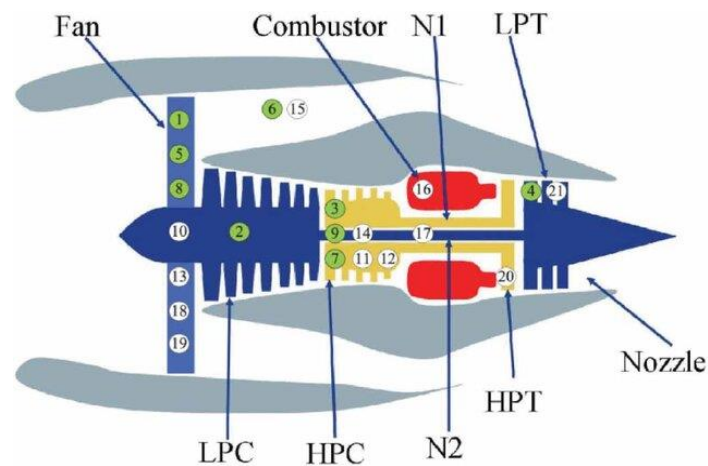

Fig. 9 C-MAPSS experimental setup [41]

## 2.9. PHM 10 CNC milling machine cutters dataset

The PHM 10 CNC milling machine cutters dataset, as shown in Fig. 10, is a valuable resource for research in the field of machine tool health monitoring and tool wear prediction. This dataset, collected during a PHM competition in 2010, features sensor data from a CNC milling machine during the machining process [42]. It provides detailed information about various cutting parameters, such as cutting force, spindle speed, feed rate, and vibration signals, as well as tool wear measurements. This dataset presents a realistic scenario of tool wear progression during machining operations, allowing researchers to develop and evaluate models for predicting tool wear, identifying the onset of tool failure, and optimizing tool replacement strategies [43]. The PHM 10 dataset has proven valuable in advancing the development of data-driven techniques, particularly those employing CNNs to analyze sensor data and predict tool wear, improving the efficiency and reliability of CNC machining processes.

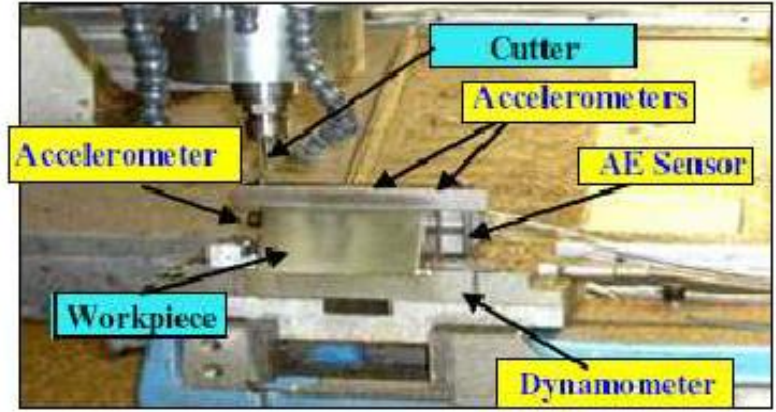

Fig. 10 CNC milling machine [44]

## 2.10. FEMTO dataset

The FEMTO dataset (also known as the PRONOSTIA dataset), collected at the FEMTO-ST Institute in France, is a valuable resource for research in bearing fault diagnosis and prognosis, as shown in Fig. 11. This dataset provides a comprehensive collection of vibration signals from various bearing conditions that have naturally occurred over time, including healthy bearings and bearings with different types and severities of faults, such as inner race, outer race, and rolling element defects. The FEMTO dataset is particularly valuable for its inclusion of multiple sensor locations, capturing vibration signals from different points on the bearing housing [45]. This allows researchers to investigate the propagation of vibration patterns and develop more accurate fault diagnosis techniques. Additionally, the dataset features varying operating conditions, including different speeds and loads, making it more realistic and applicable to real-world scenarios [46]. The availability of the FEMTO dataset has significantly contributed to the development and validation of advanced fault diagnosis techniques, particularly those utilizing CNNs to analyze vibration signals and identify bearing faults.

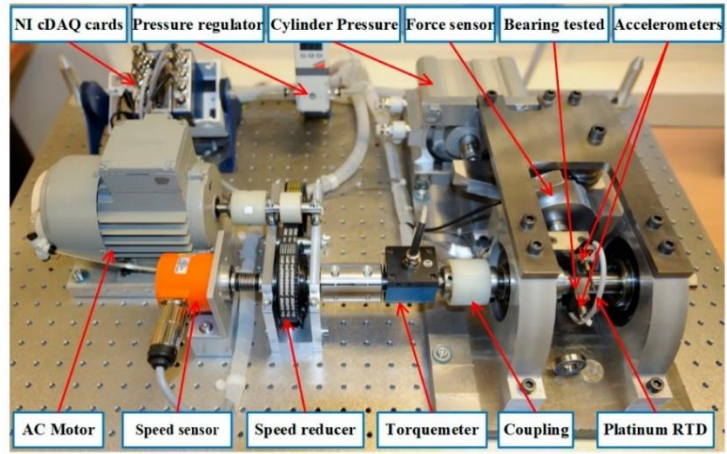

Fig. 11 PRONOSTIA experimental platform [47]

## 2.11. Wind turbine datasets

Wind turbine datasets are becoming increasingly crucial for research in condition monitoring and fault diagnosis of these renewable energy systems, as shown in Fig. 12. These datasets typically encompass a wealth of sensor readings from various components, including blade vibrations, gearbox temperatures, generator currents, and wind speed measurements. They often capture both operational data during normal operation and data associated with various fault scenarios, such as blade damage, gearbox wear, and generator malfunctions [48]. The availability of such datasets enables researchers to develop and evaluate advanced machine learning models, including those based on CNNs, to predict component degradation, detect faults early, and optimize maintenance strategies. The complexity and scale of wind turbine datasets pose unique challenges, requiring sophisticated algorithms and robust models to effectively analyze the data and provide reliable insights into wind turbine health [49]. The continued development and public sharing of these datasets will play a vital role in advancing research and innovation in wind turbine condition monitoring and predictive maintenance, contributing to the reliability and efficiency of this critical renewable energy source.

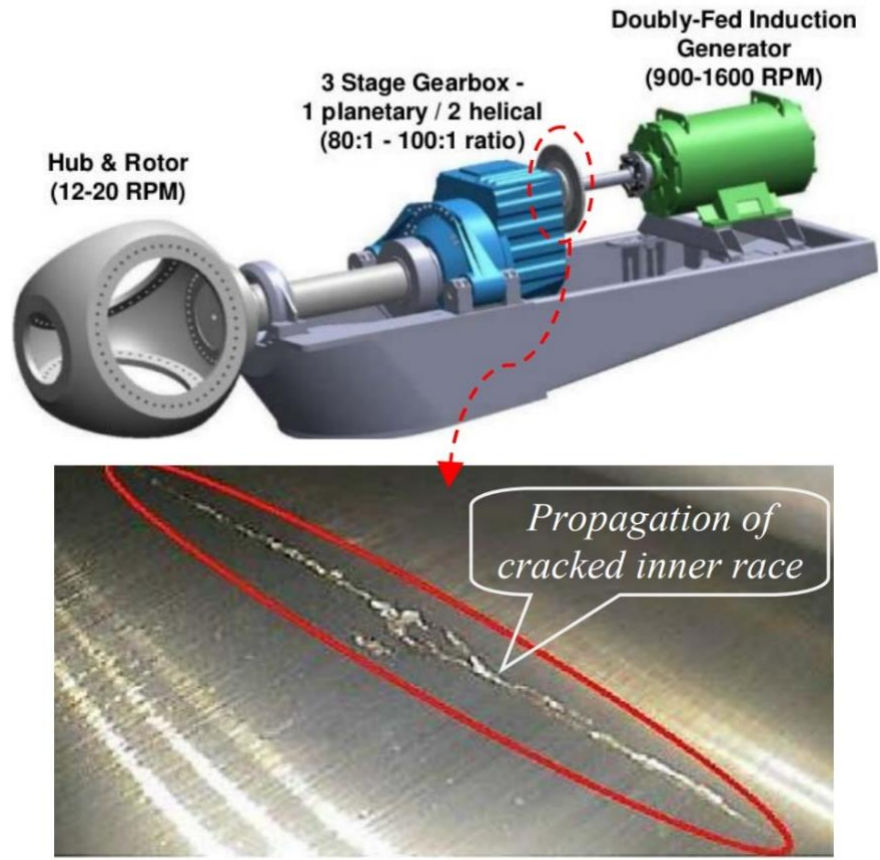

**Fig. 12** Wind turbine test rig [50]

### 2.12. Machinery failure prevention technology association (MFPT) bearing dataset

The Machinery Failure Prevention Technology Association (MFPT) bearing dataset is a valuable resource for researchers in the field of machine fault diagnosis, particularly for analyzing bearing conditions. This dataset, meticulously collected and curated by MFPT, features vibration signals from various bearing conditions, including healthy bearings and bearings with different types and severities of faults [51]. It encompasses a range of operating conditions, such as different motor speeds and applied loads, providing a realistic representation of real-world scenarios. The MFPT bearing dataset has been instrumental in advancing research in fault diagnosis, particularly in the development and validation of data-driven models, such as those based on CNNs. Researchers can use this dataset to train and test their algorithms, compare different approaches, and benchmark their performance. The availability of this publicly accessible dataset has significantly accelerated progress in the field, enabling researchers to focus on developing innovative solutions for accurate and reliable fault detection and diagnosis.

### 2.13. XJTU-SY bearing datasets

The XJTU-SY bearing datasets, as shown in Fig. 13, developed at the Xi'an Jiao tong University (XJTU) in China, provide a valuable resource for research in bearing fault diagnosis. These datasets capture vibration signals from various bearing conditions, including healthy bearings and bearings with different types and severities of faults, such as inner race, outer race, and rolling element defects [52]. The XJTU-SY datasets are notable for their inclusion of multiple operating conditions, including varying speeds and loads, which enhance their realism and applicability to real-world scenarios. The datasets also feature different bearing types, providing a broader range of data for training and testing advanced fault diagnosis techniques [53]. The availability of the XJTU-SY bearing datasets has significantly contributed to the advancement of research in this field, particularly in the development and evaluation of machine learning models, including those based on CNNs, for analyzing vibration signals and accurately identifying bearing faults.

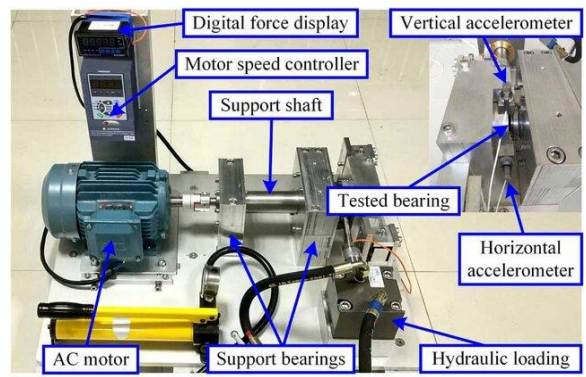

Fig. 13 XJTU-SY bearing test rig setup [54]

## 2.14. University of Connecticut gear fault dataset

The University of Connecticut gear fault dataset, as shown in Fig. 14, is a valuable resource for research in gear health monitoring and fault diagnosis. This dataset, collected at the University of Connecticut, provides a comprehensive collection of vibration signals from various gear conditions, including healthy gears and gears with different types and severities of faults, such as tooth chipping, tooth wear, and tooth breakage [55]. The dataset features varying operating conditions, including different speeds and loads, making it more realistic and applicable to real-world scenarios. The dataset's focus on gear faults makes it particularly useful for developing and evaluating advanced diagnostic techniques, including those utilizing CNNs to analyze vibration signals and identify specific gear faults [56]. The availability of this dataset has significantly contributed to the advancement of research in this field, enabling researchers to create and validate models for accurate and reliable gear health assessment.

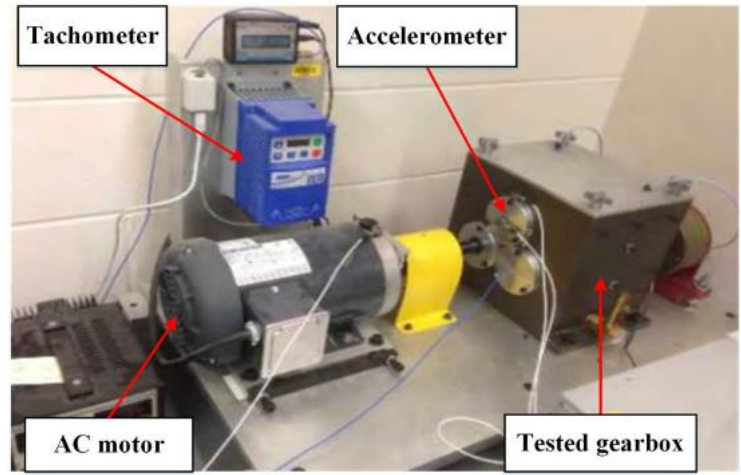

**Fig. 14** University of Connecticut experimental gear setup [57]

## 2.15. Summary

This section provided a brief overview of the data collection process for machine fault diagnosis, covering sensor selection and layout, data sampling, and data storage. This review highlights fourteen public datasets, providing descriptions and key characteristics of each.

## 3. The convolutional neural network and its variants

CNNs, a type of deep learning model, have emerged as a powerful tool in fault diagnosis, leveraging their ability to extract features and patterns from complex data. In fault diagnosis, CNNs excel at analyzing sensor data, vibration signals, or images of machinery to identify subtle anomalies indicative of impending malfunctions [58][59]. By learning hierarchical representations of these data, CNNs can detect faults even in noisy or incomplete datasets, surpassing traditional methods in accuracy and efficiency. This makes them particularly valuable in predictive maintenance, enabling early detection of faults and minimizing downtime.

### 3.1. The basic convolutional neural network

The basic convolutional neural network (CNN), as shown in Fig. 15, is a powerful tool for image recognition and other tasks involving structured data. It consists of layers that perform convolutions, a process where a small filter (kernel) slides across a matrix of input data, extracting features. These features, which can be edges, textures, or other patterns, are then passed through pooling layers, which down-sample the data while preserving important information. The resulting feature maps are then flattened and fed into fully connected layers, where they are combined to produce a final output, such as a classification or regression result [60][61]. This architecture also incorporates two common techniques, batch normalization, and dropout, to enhance model performance. Each operation will be discussed in detail in the following sections.

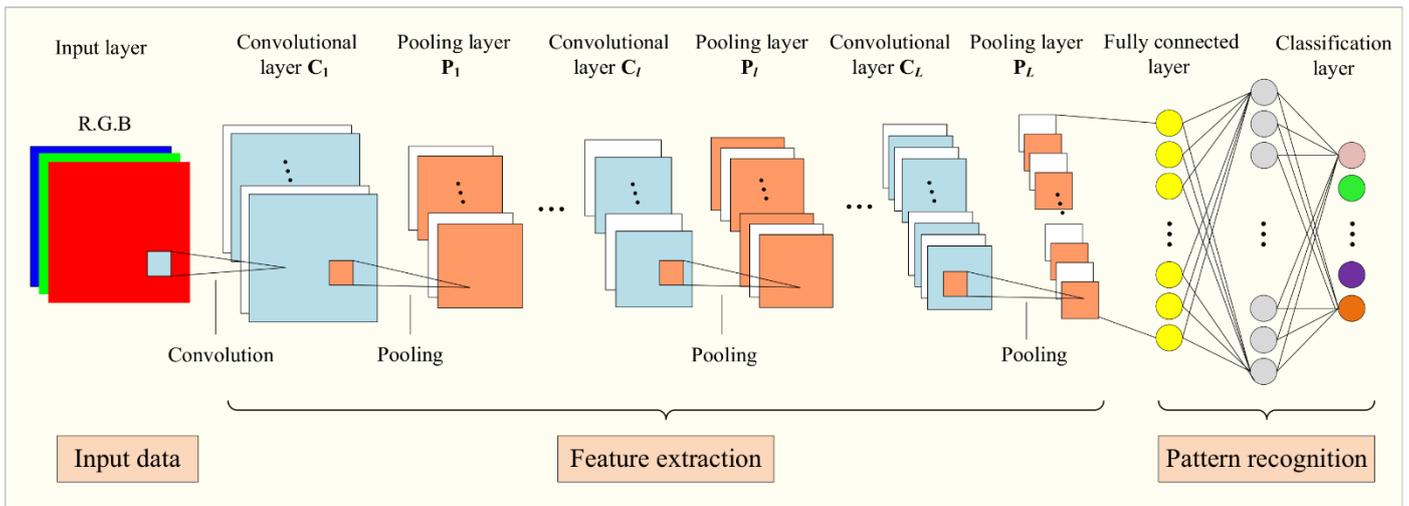

**Fig. 15** Basic structure of a CNN [62]

### 3.1.1. Convolutions

Convolutional operations play a crucial role in CNNs by extracting meaningful features from sensor data, vibration signals, or even images of machinery. These features, often subtle and hidden within the raw data, can be indicative of impending malfunctions [63]. By identifying these patterns, CNNs can detect faults before they escalate into critical failures, enabling proactive maintenance and minimizing downtime. In fault diagnosis, convolutions operate as follows:

**1. Input Data:** The input data consists of time-series signals from sensors, processed frequency domain information, time-frequency images, or images capturing machine operations. These signals often contain intricate patterns, some subtle, that indicate a deviation from normal behavior.

**2. Filters (Kernels):** Filters in fault diagnosis are designed to capture specific patterns related to various faults. These filters might focus on:

- **Frequency Components:** Identifying specific frequencies in vibration signals that are associated with certain faults.
- **Temporal Trends:** Detecting sudden changes or drifts in sensor readings that indicate anomalies.
- **Spatial Features:** Recognizing unusual patterns in images of machinery, such as cracks, wear, or misalignment.

**3. Convolution Process:** The filter slides across the input signal, performing element-wise multiplications and summations. The result is a new signal, the feature map, where each point represents the strength of the filter's pattern in the original signal.

**4. Feature Extraction:** Through convolutional operations, the CNN extracts features that might be missed by traditional analysis methods. These features represent the "signature" of specific faults, allowing the model to detect them even in noisy or complex data.

The convolution enables the extraction of complex features from noisy data, leading to more accurate fault detection and classification.

### 3.1.2. Activation function

Activation functions play a crucial role in CNNs for fault diagnosis by introducing non-linearity into the model, allowing it to learn complex relationships within the data. In essence, they act as "decision-makers" within the network, deciding which features are important and how they contribute to the final diagnosis [64]. In fault diagnosis, activation functions operate as follows:

- The input to an activation function is the output of a convolutional layer or a fully connected layer. This output represents the extracted features, typically a set of numerical values.
- The activation function then applies a non-linear transformation to these input values. This transformation introduces non-linearity into the model, enabling it to capture complex relationships between features that might not be linearly separable.
- The output of the activation function is a transformed set of values representing the "activated" features. These activated features are then passed on to the next layer of the network.

The most common activation functions that are used by researchers include:

1. **ReLU (Rectified Linear Unit)**: ReLU is a widely used activation function with CNNs for fault diagnosis. It introduces non-linearity by setting any negative values to zero and leaving positive values unchanged. This helps prevent vanishing gradients during training and allows the model to learn features more efficiently.
2. **Sigmoid**: The sigmoid function squashes the input values between 0 and 1, making it suitable for tasks involving binary classification (e.g., classifying a sensor reading as "normal" or "faulty").

3. **Tanh (Hyperbolic Tangent)**: Like sigmoid, the tanh function also outputs values between -0 and 1. It is often preferred over sigmoid due to its centered output range, which can improve learning in some cases.

In conclusion, activation functions are crucial elements in CNNs for fault diagnosis, enabling the model to learn complex relationships, select relevant features, and effectively differentiate between normal and faulty states. They are a key component of the decision-making process within the network, allowing for accurate and efficient fault diagnosis.

### 3.1.3. Pooling

Pooling in convolutional neural networks (CNNs), as shown in Fig. 16, plays a crucial role in fault diagnosis by simplifying the feature maps generated by convolutional layers, while preserving essential information for accurate diagnosis [65]. This simplification process, often referred to as down sampling, helps the model become more robust to variations in the input data and improves its efficiency by reducing the number of parameters. In CNNs, pooling works as follows:

1. The input to a pooling layer is a feature map generated by a convolutional layer. This map represents the extracted features, highlighting patterns relevant to potential faults.
2. A pooling operation is applied to the feature map, typically using a small window (e.g., 2x2). Common pooling operations include:
   - **Max Pooling:** This selects the maximum value within each window, effectively capturing the strongest feature.
   - **Average Pooling:** This calculates the average value within each window, providing a smoother representation of the features.
3. **Down sampling:** The pooling operation down samples the feature map, reducing its spatial dimensions. This process reduces the number of parameters the model needs to learn, improving efficiency and preventing overfitting.
4. **Output:** The output of the pooling layer is a down-sampled feature map, retaining the essential information for fault diagnosis while reducing noise and redundancy.

In conclusion, pooling in CNNs is a powerful technique that simplifies feature maps, reduces computational complexity, and enhances the model's robustness to variations in the input data. By preserving essential information, while discarding irrelevant details, pooling contributes to accurate and efficient fault detection and diagnosis.

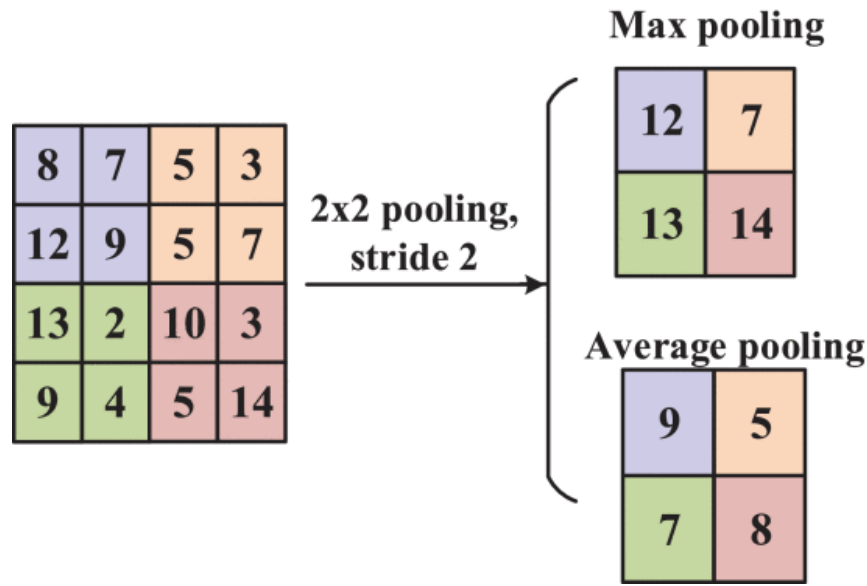

Fig. 16 Basic procedure of pooling [66]

### 3.1.4. Batch normalization

Batch normalization, as shown in Fig. 17, a powerful technique often incorporated into CNNs, plays a crucial role in enhancing the performance of fault diagnosis models by addressing the issue of internal covariate shifts. This shift occurs when the distribution of activations within the network changes during training, leading to slower convergence and instability [67][68]. Batch normalization helps to stabilize the training process, leading to faster convergence and improved accuracy. Here's how batch normalization works in the context of fault diagnosis:

- The input to the batch normalization layer is the activations of a previous layer, typically a convolutional or fully connected layer. These activations represent the extracted features from the input data.
- The batch normalization layer normalizes the input activations by calculating the mean and standard deviation of the activations within the current batch. This helps to estimate the distribution of activations for the current batch. Normalizing the activations using the calculated mean and standard deviation (i.e., transforms the activations to have zero mean and unit variance).
- After normalization, the activations are scaled and shifted using learnable parameters (gamma and beta). This allows the model to retain the expressiveness of the activations while ensuring they are within a suitable range for the subsequent layers.

Batch normalization addresses the problem of internal covariate shift by ensuring that the distribution of activations remains relatively stable during training. This leads to faster convergence and reduces the risk of the model getting stuck in a local minima. By reducing the internal covariate shift, batch normalization helps the model generalize better to unseen data. This is crucial for fault diagnosis, where the model needs to reliably identify faults in real-world scenarios. It also acts as a regularizer, preventing the model from overfitting to the training data. This is particularly beneficial for fault diagnosis, where limited data is often available. It also helps to improve the flow of gradients during training, allowing the model to learn effectively and adjust its parameters more efficiently. Batch normalization helps to mitigate the issue of vanishing or exploding gradients, particularly in deep networks. This allows the model to train deeper architectures with fewer issues.

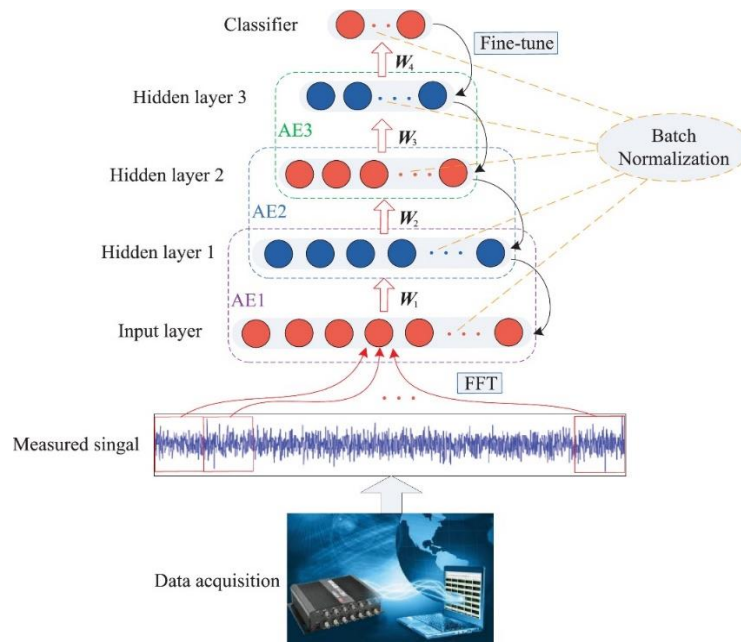

Fig. 17 Batch normalization architecture [69]

In conclusion, batch normalization is a powerful tool in CNNs for fault diagnosis, enhancing the stability, efficiency, and generalizability of the model. By addressing the issue of internal covariate shift, batch normalization significantly contributes to improved accuracy and reliability in diagnosing faults.

### 3.1.5. Dropout

Dropout, a regularization technique commonly used in CNNs, plays a crucial role in enhancing the robustness and generalizability of fault diagnosis models by preventing overfitting. Overfitting occurs when a model learns the training data too well, leading to poor performance on unseen data [70]. Dropout helps address this issue by randomly dropping out neurons during training, forcing the network to learn more robust and generalizable features. For fault diagnosis, dropout works as follows:

- During each training iteration, dropout randomly deactivates a certain percentage of neurons in a layer, effectively removing them from the network's computation. This creates a smaller, randomly thinned network.
- By randomly dropping neurons, dropout prevents neurons from becoming too dependent on each other. This encourages the network to learn more independent and robust features.
- Dropout can be viewed as creating an ensemble of multiple smaller networks, each with a different set of active neurons. This ensemble effect improves the model's generalizability by averaging the predictions of these smaller networks.

Dropout reduces the model's reliance on specific features, making it less prone to overfitting. This is crucial in fault diagnosis, where limited data availability can lead to overfitting and poor generalization. By forcing the network to learn more robust and generalizable features, dropout improves the model's performance in the presence of noise or variations in the input data. This is essential for reliable fault detection in real-world scenarios. Dropout helps the model generalize better to unseen data. This is crucial for fault diagnosis, where the model needs to be able to accurately detect faults in previously unseen cases. Dropout is relatively simple to implement and can be easily incorporated into existing CNN architectures.

In conclusion, dropout is a valuable regularization technique in CNNs for fault diagnosis. By preventing overfitting and encouraging the network to learn more robust features, dropout helps to improve the model's generalizability, leading to more reliable and accurate fault detection.

### 3.1.6. Fully-connected layers

Fully-connected layers are another crucial component in CNNs for fault diagnosis by combining extracted features from convolutional layers into a meaningful representation suitable for making final predictions [71][72]. These layers act as the decision-making engine, taking the high-level features learned by the convolutional layers and transforming them into a form that can be used to classify or predict faults. Fully-connected layers work in the following manner:

- The input to a fully-connected layer is the output of the previous layer, typically a pooling layer or the final convolutional layer. This output represents the extracted features from the input data, often in the form of a flattened vector.
- In a fully-connected layer, each neuron receives an input from all neurons in the previous layer. This means that every feature from the previous layer contributes to the calculation of the output of each neuron in the fully-connected layer.

- The fully-connected layer performs a linear transformation on the input features by multiplying them with weights and adding biases. These weights are learned during training to optimize the network's performance.
- After the linear transformation, an activation function (e.g., ReLU, sigmoid) is applied to introduce non-linearity, allowing the network to learn complex relationships between features.
- The output of the fully-connected layer is a set of values representing the model's predictions. The number of neurons in the output layer corresponds to the number of classes in the fault diagnosis task.

**Role of Fully-Connected Layers in Fault Diagnosis:**

- **Feature Combination:** Fully-connected layers combine the high-level features extracted by the convolutional layers into a comprehensive representation of the input data. These features can represent various aspects of the machine's operation, such as vibration patterns, sensor readings, or visual cues.

- **Decision-Making:** The fully-connected layers act as the decision-making component of the CNN. They take the combined features and learn relationships between them to predict the presence or absence of specific faults.

- **Classification or Regression:** Depending on the fault diagnosis task, fully-connected layers can be configured for either classification or regression:

  - **Classification:** The output layer will have neurons representing different fault classes (e.g., bearing failure, pump malfunction, etc.).

  - **Regression:** The output layer will predict a continuous value representing the severity of a fault or the remaining useful life of a component.

In conclusion, fully-connected layers play a critical role in fault diagnosis by integrating the high-level features learned via convolutional layers, enabling the CNN to make final predictions about the presence, type, or severity of faults. They act as the decision-making engine, leveraging the extracted features to provide accurate and insightful diagnoses.

### 3.1.7. Decision layer

The decision layer, often the final layer in a CNN for fault diagnosis, is where the model makes its final prediction about the presence, type, or severity of a fault based on the processed features. It acts as the "judge" of the network, taking the combined knowledge from previous layers and transforming it into a clear and actionable output [73]. The decision layer mechanism is as follows:

1. The input to the decision layer is the output of the last fully connected layer. This output typically represents a vector of values, each corresponding to a specific fault class or a continuous value representing the severity of a fault.

2. The decision layer implements a specific mechanism to determine the final output based on the input values:

- **For classification**: In a multi-class classification task, the decision layer might use a SoftMax function to convert the output values into probabilities for each fault class. The class with the highest probability is then selected as the predicted fault.
- **For regression**: In a regression task, the decision layer directly outputs a continuous value representing the predicted fault severity, remaining useful life, or other relevant metric.

3. The output of the decision layer is the final prediction of the CNN model, providing a clear indication of the fault status or severity.

In general, there are two types of decision layers that have been used by the researchers:

- **SoftMax Layer**: Used for multi-class classification, the softmax layer converts the output values into probabilities that sum to 1. This allows the model to assign a confidence score to each fault class.
- **Linear Layer**: Used for regression tasks, a linear layer outputs a continuous value representing the predicted fault severity or other relevant metric.

In conclusion, the decision layer in a CNN for fault diagnosis acts as the final stage, transforming the processed information into a clear and actionable prediction. By using appropriate decision-making mechanisms, the decision layer enables the CNN to provide reliable and informative diagnoses, facilitating effective fault management and maintenance strategies.

### 3.1.8. Optimizer

In the context of fault diagnosis using CNNs, the optimizer plays a crucial role in guiding the learning process, ensuring the model effectively learns to identify and classify faults from the provided data [74]. It acts as the "trainer," adjusting the model's parameters (weights and biases) during training to minimize errors and improve performance. The procedure of implementing the optimizer in a CNN is achieved through the following steps:

1. **Objective Function**: The objective function, also known as the loss function, measures the difference between the model's predictions and the actual fault labels. The goal is to minimize this loss function, indicating a better fit to the data.

2. **Parameter Updates**: The optimizer calculates the gradients of the loss function with respect to the model's parameters. These gradients represent the direction and magnitude of change needed to minimize the loss.

3. **Gradient Descent**: The optimizer uses these gradients to update the model's parameters iteratively, moving them in the direction of decreasing loss. Different optimizers employ different strategies for updating the parameters, leading to varying performance characteristics.

Common optimizers that have been used by researchers for fault diagnosis include:

- **Stochastic Gradient Descent (SGD):** SGD is a basic and widely used optimizer. It updates parameters based on the gradients calculated from a single randomly selected batch of training data. While simple, SGD can be slow to converge and may get stuck in a local minima.

- **RMSprop**: RMSprop, similar to AdaGrad, uses a moving average of squared gradients to adjust the learning rates for each parameter. This helps to prevent oscillations in the learning process and enables faster convergence.

- **Adam**: Adam combines the advantages of AdaGrad and RMSprop, adapting learning rates for each parameter based on their historical gradients. This adaptive learning rate strategy generally leads to faster convergence and better performance.

In conclusion, the optimizer is a critical component in training CNNs for fault diagnosis, effectively guiding the learning process to minimize errors and enhance performance. By selecting and configuring an appropriate optimizer, models can converge efficiently and provide accurate diagnoses of various faults in real-world settings.

### 4. Advancements in convolutional network architectures

CNNs have evolved significantly since their inception, leading to various variants and extensions that enhance their capabilities and applicability across diverse domains. These advancements address specific challenges, improve performance, and enable CNNs to tackle increasingly complex tasks. Notable variants and extensions are discussed in the following sections.

## 4.1. Depth-wise Separable Convolutions (DSCs)

Depth-wise separable convolutions (DSCs), as shown in Fig. 18, is a powerful technique in deep learning that offers significant advantages in terms of computational efficiency and model size reduction. This makes it particularly valuable for applications with limited computational resources and memory, like fault diagnosis in industrial settings [75]. For feature extraction, DSCs use two techniques, namely depth-wise convolutions and point-wise convolutions. Depth-wise convolutions apply separate filters to each input channel, allowing for efficient feature extraction tailored to the specific characteristics of each channel [76]. This is crucial in fault diagnosis as different features might indicate different types of faults. On the other hand, point-wise convolutions combine extracted features by applying a single convolution across all channels. This enables the network to learn complex relationships between features, facilitating accurate fault classification.

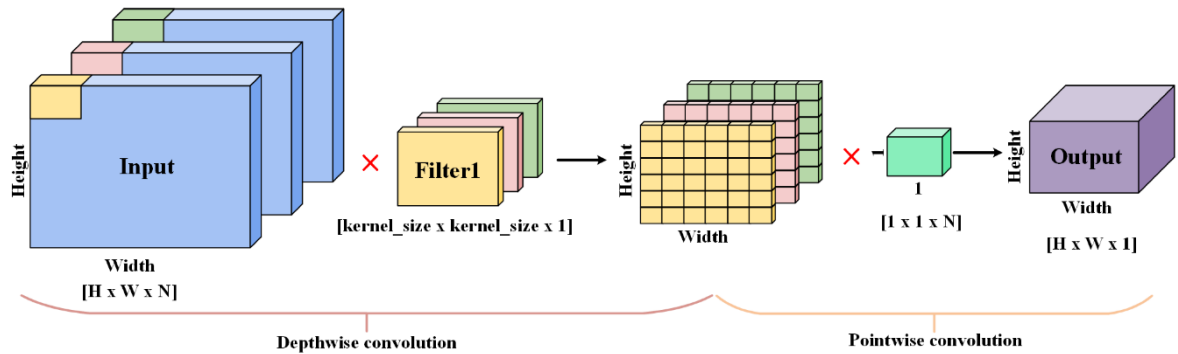

Fig. 18 Architecture of DSCs [77]

The DSCs technique requires significantly fewer parameters compared to traditional convolutions, leading to reduced computation time and memory usage. This is especially beneficial for real-time fault diagnosis systems with limited processing power. The smaller number of parameters translates to a smaller model size, making it easier to deploy and manage on resource-constrained devices like edge computers. Despite the reduced complexity, DSCs can achieve comparable or even better accuracy than traditional convolutions in fault diagnosis applications. This is because it allows the network to learn more specialized and efficient features. DSCs are known to be more robust to variations in data, which is crucial in fault diagnosis, where data can be noisy and inconsistent.

DSCs offer several advantages for fault diagnosis applications, including reduced computational complexity, improved accuracy, and increased robustness. Its ability to extract relevant features, while minimizing model size, makes it a powerful tool for building efficient and reliable fault diagnosis systems in diverse industrial settings

## 4.2. Dilated convolutions

Dilated convolutions, also known as "atrous convolutions", are a powerful technique in deep learning that allows for capturing long-range dependencies in data without increasing the number of parameters or computational complexity, as shown in Fig. 19. This makes it particularly valuable for fault diagnosis applications where understanding the context of the data is crucial for accurate identification of faults [78]. Dilated convolutions allow the network to access information from a wider range of surrounding data points, effectively increasing the receptive field. This is crucial for fault diagnosis as faults often manifest as subtle deviations in a larger context, not just isolated anomalies. By varying the dilation rate, dilated convolutions can extract features at different scales, capturing both local and global patterns in the data [79]. This enables the network to identify faults with varying degrees of complexity and duration.

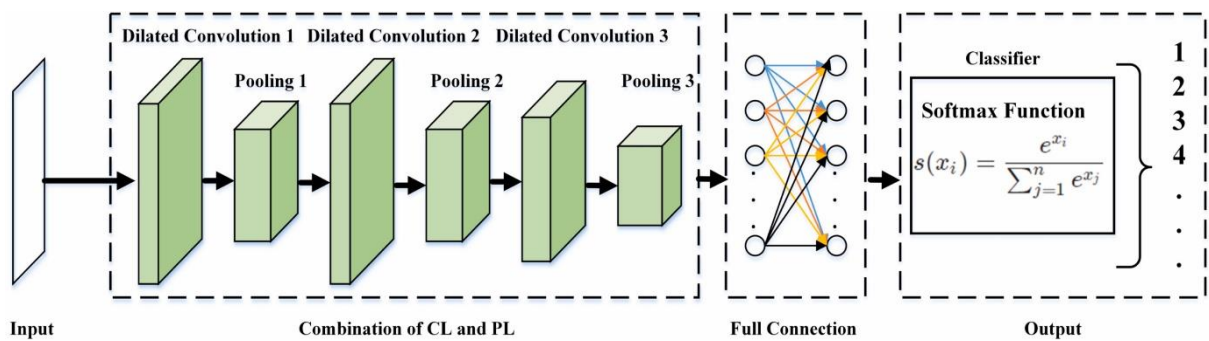

Fig. 19 Architecture of dilated convolutions [80]

Dilated convolutions empower fault diagnosis systems to extract contextual information from data, capturing long-range dependencies and enabling the identification of subtle or complex faults [81]. Their ability to analyze data at multiple scales and improve accuracy with minimal computational overhead makes them a valuable tool for developing advanced fault diagnosis systems in various industries.

## 4.3. Residual networks

Residual networks (ResNets), as shown in Fig. 20, are a powerful class of deep neural networks that have revolutionized the field of image recognition and have also proven highly effective in various other domains, including fault diagnosis. ResNets introduce "skip connections", which allow information to directly pass through multiple layers without any transformation. This helps to overcome the vanishing and exploding gradient problem that can occur in very deep networks, enabling the training of deeper architectures and extracting more complex features [82]. Skip connections facilitate the flow of gradients through the network, making it easier to optimize the model and achieve better performance. Instead of learning the complete mapping function, ResNets learn the "residual function" - the difference between the desired output and the input. This makes the optimization process more stable and enables the network to learn intricate patterns more effectively. ResNets tend to generalize better to unseen data, reducing the risk of overfitting and improving the robustness of the fault diagnosis system.

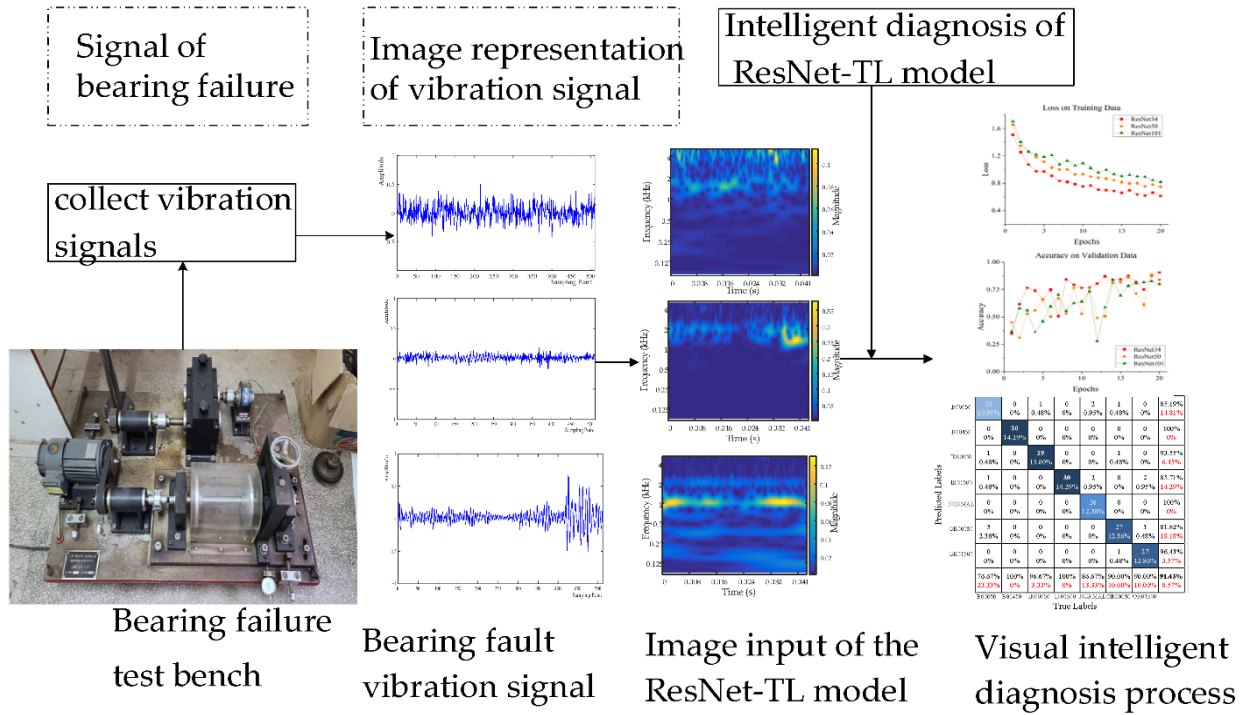

Fig. 20 Architecture of a residual network [83]

Residual networks offer significant advantages for fault diagnosis applications. Their ability to overcome gradient issues, enhance accuracy, and improve generalization makes them a powerful tool for building robust and reliable fault diagnosis systems [84]. ResNets can effectively analyze sensor data, identify anomalies, classify faults, and contribute to predictive maintenance strategies in various industrial settings.

### 4.4. Squeeze-and-excitation (SE) networks

SE networks, as shown in Fig. 21, are a powerful technique that enhance deep learning models by allowing them to selectively focus on relevant features in the input data. This selective attention mechanism can greatly improve the performance of fault diagnosis systems, particularly in scenarios where subtle patterns and complex relationships between features are crucial for the accurate identification of faults [85]. The squeeze step of this network aggregates spatial information from feature maps, creating a compact representation of the overall feature distribution. This step captures the global context of the features and their relative importance. The excitation step uses a fully connected neural network to generate channel-wise weights, effectively recalibrating the importance of each feature channel based on its contribution to the overall prediction [86]. The final step, reweighting, generates the weights that are then applied to the original feature maps, selectively enhancing the contribution of important channels and suppressing less relevant ones.

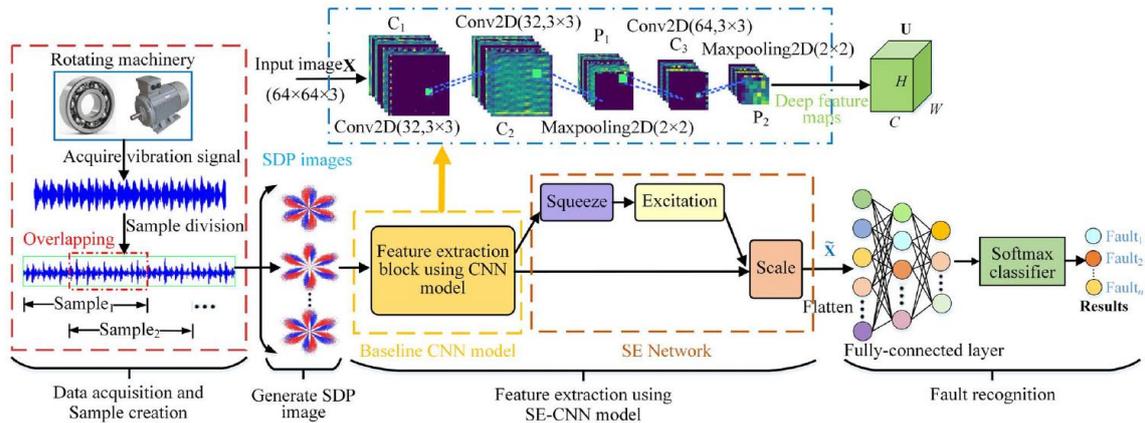

**Fig. 21** Architecture of a SE network [87]

SE networks offer a powerful way to enhance deep learning models for fault diagnosis applications. By selectively focusing on relevant features, they can improve accuracy, robustness, and interpretability, making them a valuable tool for developing more effective and reliable fault diagnosis systems in various industrial settings.

### 4.5. Recurrent convolutional neural networks (RCNNs)

RCNNs, as shown in Fig. 22, combine the strengths of CNNs and recurrent neural networks (RNNs) to effectively analyze time-series data, making them particularly well-suited for fault diagnosis in industrial settings. RCNNs utilize convolutional layers to extract spatial features from the input data, capturing local patterns and relationships within each time step. The extracted features are then fed into recurrent layers, such as long short-term memory (LSTM) networks or gated recurrent units (GRUs), which are designed to capture temporal dependencies and learn long-term patterns over time [88]. By considering both spatial and temporal aspects of the data, RCNNs can achieve higher accuracy in fault classification compared to models that only analyze spatial features or temporal patterns individually. RCNNs can learn to detect subtle changes in the data that indicate early signs of faults, allowing for proactive maintenance and preventing costly failures. RCNNs can effectively model complex fault patterns that involve both spatial and temporal variations, allowing for accurate diagnosis of intricate failure mechanisms.

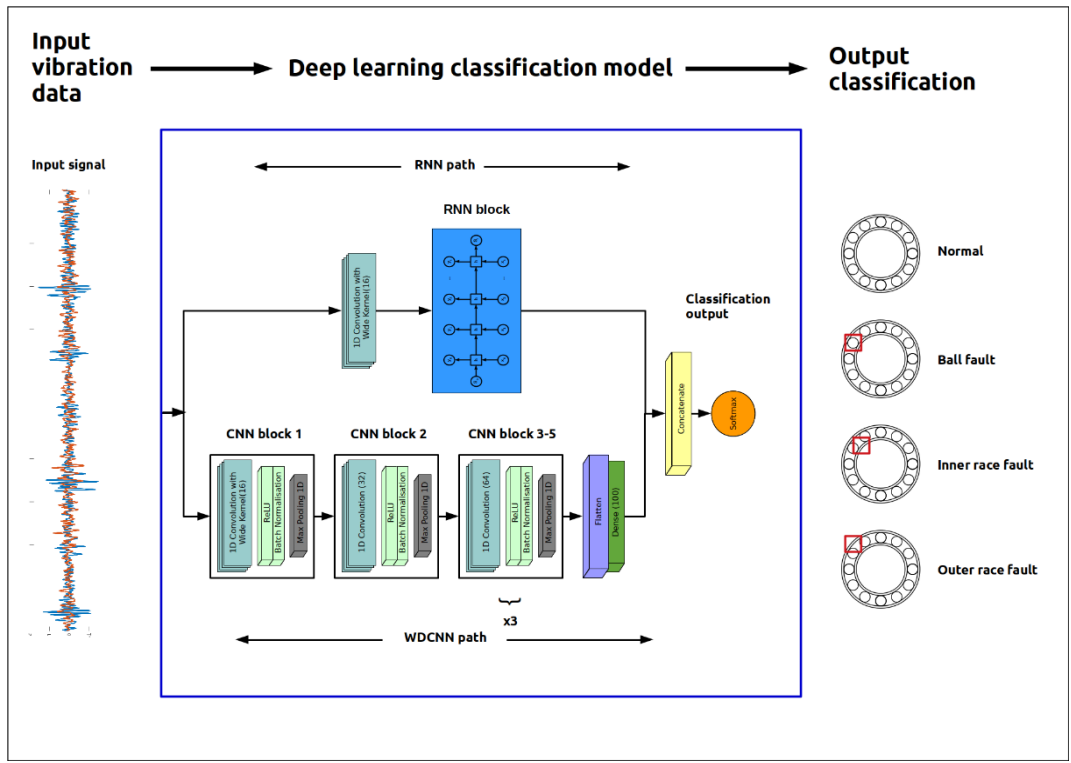

**Fig. 22** Architecture of RCNNs [89]

### 4.6. Densely connected convolutional networks (DenseNets)

DenseNets are a powerful architecture, as shown in Fig. 23, that significantly enhance feature propagation and learning in deep convolutional networks. This makes them particularly beneficial for fault diagnosis, where extracting and combining rich features from data is crucial for accurate and robust fault identification. DenseNets connect each layer to all preceding layers, allowing features from earlier layers to be directly propagated to subsequent layers [90]. This creates a "feed-forward" network that efficiently combines information from multiple layers. The dense connections encourage the reuse of features across the network, promoting the learning of more discriminative and informative representations. By leveraging the collective knowledge of all preceding layers, DenseNets can achieve higher accuracy in fault classification, particularly when dealing with complex fault patterns. The dense connections mitigate the vanishing gradient problem, ensuring that important features are preserved and propagated throughout the network [91]. DenseNets are more robust to noise and variations in data, as the dense connections allow the network to learn redundant features and compensate for noisy or missing information.

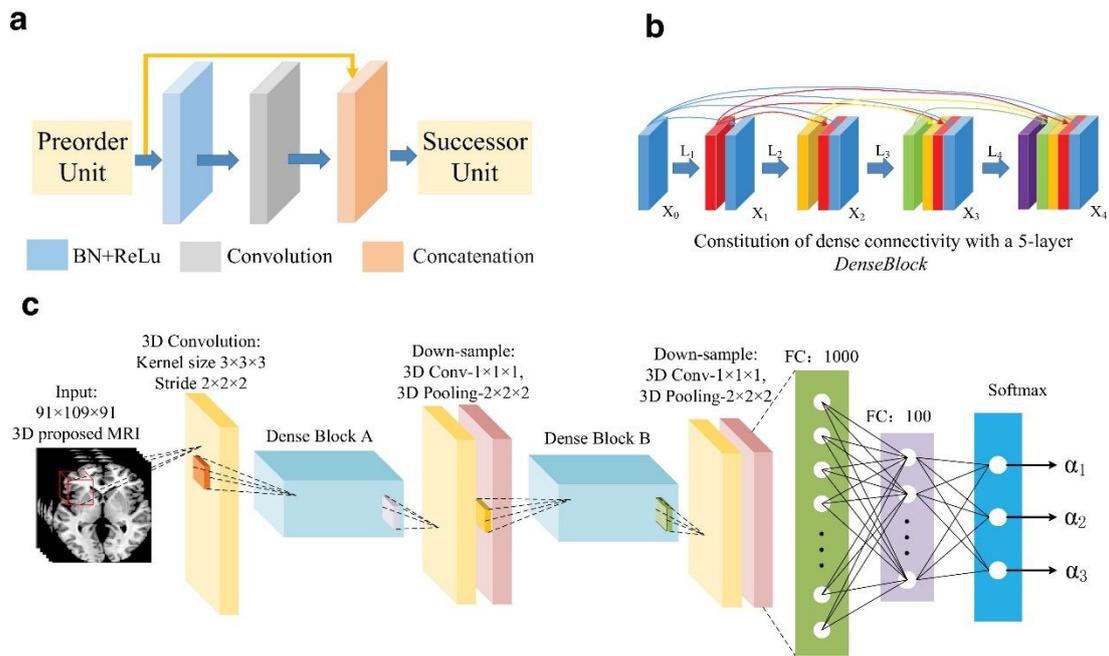

**Fig. 23** Architecture of DenseNets [92]

DenseNets offers a powerful architecture for fault diagnosis, enabling more efficient feature propagation and richer feature learning. Their ability to combine information from multiple layers, reduce feature loss, and enhance robustness makes them a valuable tool for building accurate and reliable fault diagnosis systems in various industrial settings.

### 4.7. Generative adversarial convolutional networks (GACNs)

GACNs, as shown in Fig. 24, are a powerful deep learning technique that combine the strengths of generative adversarial networks (GANs) and CNNs for generating realistic and diverse synthetic data. This makes them particularly valuable for fault diagnosis applications, where obtaining sufficient labeled data for training is often challenging [93]. The generator component of a GACN learns the distribution of normal data, capturing the underlying patterns and characteristics of healthy machine operation. The discriminator component distinguishes between real data (normal operation) and synthetic data generated by the generator [94]. By analyzing the differences between real and synthetic data, GACNs can identify anomalies or deviations from normal operating conditions, indicating potential faults. GACNs can generate realistic synthetic data that closely resembles the characteristics of real data, even for rare or uncommon fault scenarios [95]. This synthetic data can be used to augment the training dataset, improving the model's robustness and generalization capabilities. GACNs can address data imbalance problems by generating more samples of rare fault types, ensuring that the model learns to identify them accurately.

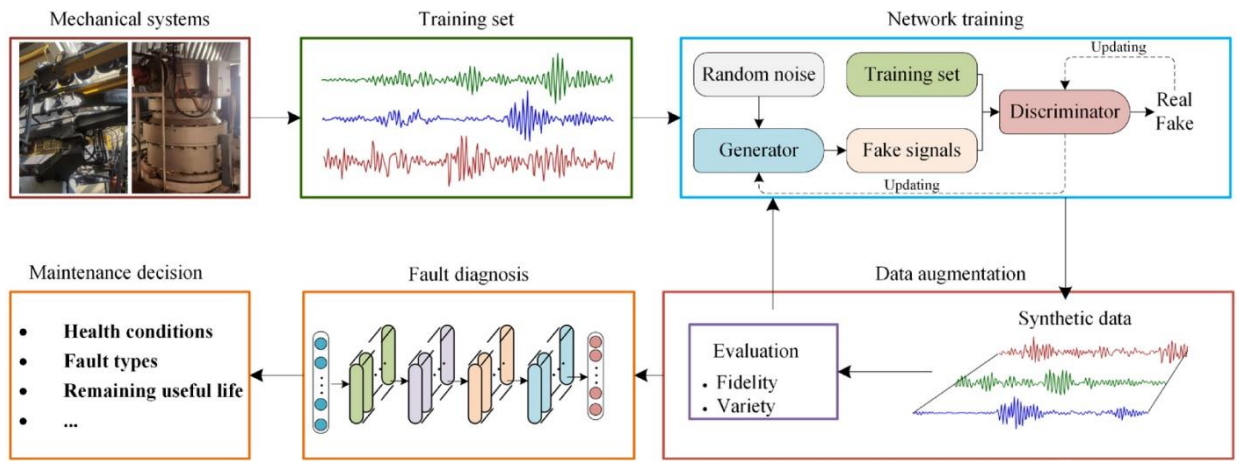

Fig. 24 Architecture of GACNs [96]

GACNs offer a powerful approach to fault diagnosis, particularly in scenarios where data is limited or imbalanced. Their ability to generate realistic synthetic data enhances anomaly detection capabilities, improves model robustness, and enables more effective training, making them a valuable tool for building intelligent and reliable fault diagnosis systems.

## 5. Progress of CNN development in machine fault diagnosis

Fault classification, directly inspired by image classification techniques, is the oldest and most extensively researched area within CNFD. This section offers a systematic review of applications within this field. To improve clarity and organization, studies are categorized based on the convolutional network architecture employed. This results in three distinct groups: 2D convolutional network-based classification, 1D convolutional network-based classification, and fault classification leveraging variants of convolutional networks.

Early applications of CNNs in machine fault diagnosis directly adopted the original 2D structure, replicating image processing techniques. However, as mechanical signals are predominantly 1D time series, a key challenge arose in converting 1D data into a 2D format suitable for 2D CNNs. Researchers have developed various signal processing techniques which can be used for converting 1D data into a 2D format. Data matrix transformation is a technique that involves directly arranging raw mechanical data into a 2D format for input into a CNN. Several researchers have implemented this technique. For instance, Xiong et al. [97] addresses the limitations of traditional dimensionless index-based methods for bearing fault diagnosis in petrochemical systems. The proposed approach combines mutual dimensionless theory and similar Gram matrix pre-processing with a CNN. Wen et al. [98] proposed a novel CNN-based approach that eliminates the need for handcrafted features by converting signals into 2D images and leveraging the feature extraction capabilities of CNNs. Li et al. [99] address the limitations of traditional CNN-based bearing fault diagnosis methods, which often require manual data conversion and expert knowledge to extract features from 1D sensor data. The proposed ST-CNN method overcomes these challenges by integrating the S-transform (ST) algorithm into the CNN architecture. Li et al. [100] introduced a novel feature extraction scheme for roller bearing fault diagnosis, combining a generalized S transform (GST) with two-dimensional non-negative matrix factorization (2DNMF). Yi et al. [101] proposed an approach using an optimized relative transformation matrix (RTM) with a bacterial foraging algorithm (BFA-ORTM) to address the challenges of fault diagnosis in aluminium electrolysis cells, which involve extracting meaningful features from complex and highly coupled data. Zhao et al. [102] combined the S transform, gray level co-occurrence matrix (GLCM), and multi-class support vector machine (Multi-SVM) to provide a robust and accurate method for identifying faults, contributing to enhanced reliability and maintenance in machinery applications. Yang et al. [103] demonstrate the potential of 1D-CNNs for efficient and accurate fault diagnosis in analog circuits. The method's simplicity, effectiveness, and ability to handle raw data without manual feature engineering highlight its significant contribution to the field of analog circuit diagnostics. Ince et al. [104] presented a promising solution for real-time motor fault detection by integrating a 1D-CNN into a single efficient learning system. This approach holds the potential to improve the reliability and safety of motor systems by enabling early fault detection and proactive maintenance. Zhang et al. [105] presented a robust and intelligent approach to bearing fault diagnosis, showcasing the potential of combining optimization algorithms with deep learning techniques. The enhanced signal processing capabilities and improved fault classification accuracy make this method a valuable tool for improving the reliability and safety of machinery with roller bearings. Gao et al. [106] presented a novel and effective approach to fault diagnosis in analog circuits. The combination of advanced optimization algorithms with a 1D-CNN and an attention mechanism demonstrates the potential to overcome the challenges associated with complex circuit analysis, leading to improved fault identification and enhanced system reliability. Sahu et al. [107] underscore the effectiveness of multi-input 1D-CNNs for accurate and reliable bearing fault diagnosis. The research contributes to the development of robust and data-driven solutions for improving the reliability and safety of rotating machinery. Saufi et al. [108] addressed the challenges of roller bearing fault diagnosis under low operating speeds, where traditional methods struggle due to minimal fault-related signal information. The proposed approach utilizes data fusion from acoustic emissions and vibration sensors with a novel 1D-CNN optimized by a differential evolution algorithm (DE-1D-CNN).

Image transformation is another technique that involves converting 1D mechanical signals into pixel-based images. This approach allows researchers to leverage the strengths of 2D CNNs, originally designed for image processing, to analyze mechanical data. For instance, Alshorman et al. [109] significantly contributed to the field of induction motor health monitoring by providing a comprehensive overview of image-based intelligent techniques. They highlight the potential for improved accuracy, early fault detection, and proactive maintenance strategies, ultimately contributing to the reliability and efficiency of induction motors in industrial operations. Luczak et al. [110] presented a novel approach for fault diagnosis in electromechanical systems, leveraging time-frequency analysis with CNNs to achieve efficient and accurate fault identification. Zhang et al. [111] address the challenge of accurately diagnosing gearbox faults in quayside container cranes (QCCs), presenting a novel approach that combines a frequency-domain Markov transformation field (FDMTF) with a multi-branch residual convolutional neural network (MBRCNN). Lian et al. [112] presented an approach that integrates cross-domain fusion images and a lightweight feature-enhanced network for fault diagnosis. Temporal and frequency are used to transform time series information into images. Fan et al. [113] integrated adaptive batch normalization with a CNN, which effectively extracts features that remain stable despite variations in load conditions. These studies showcase the effectiveness of image transformations in converting 1D mechanical data into a format suitable for 2D CNNs, leading to advancements in fault diagnosis.

Another approach to converting 1D mechanical signals for CNN inputs involves utilizing statistical features extracted from the time or frequency domains. Chen et al. [114] used statistical metrics derived from vibration signals in both the time and frequency domains as inputs for their model. Shirazi et al. [115] combined feature extraction and damage prediction into a single step, reducing complexity and improving efficiency in structural health monitoring. Ahmadzadeh et al. [116] integrated a time-distributed 1D CNN with LSTM layers, which enables the effective extraction of both spatial and temporal features from raw multi-sensor time histories. Pang et al. [117] proposed an explainable and lightweight 1-D CNN (ELCNN) model based on square global average pooling (S-GAP) and improved vibration signals to address the challenges of high computational complexity and poor interpretability. Kaya et al. [118] introduced an innovative fault diagnosis approach that includes a feature extraction technique that utilizes the MM-1D-LBP technique in tandem with a hybrid deep learning network, combining LSTM and 1D-CNN architectures. These studies demonstrate the versatility of extracting statistical features from the time and frequency domains as inputs for CNNs, leading to successful applications in fault diagnosis.

Wavelet transform (WT) pre-processing converts mechanical time series data into 2D time-frequency (TF) representations for use as a network input. Yang et al. [119] highlight the progression and significance of time-frequency analysis (TFA), which maps signals onto the time-frequency plane. While traditional non-parameterized TFA can theoretically analyze any signal, it falls short for complex signals. Parameterized TFAs, in contrast, offer superior representations by introducing kernel functions with additional parameters. Santaniello and Russo [120] introduced a novel approach for multiclass damage detection utilizing acceleration responses in conjunction with the synchrosqueezing transform (SST) and deep learning algorithms. The proposed methodology effectively classifies time-series data representing accelerometer responses of a bridge, identifying different types of damage scenarios. Kumar et al. [121] presented a fault identification scheme for the direct-shift gearbox utilizing variational mode decomposition (VMD) and a CNN. Vashishtha and Kumar [122] proposed a deep learning-based scheme utilizing time-varying filter-based empirical mode decomposition (TVF-EMD), an amended grey wolf optimization (AGWO) algorithm and a CNN. These studies demonstrate the effectiveness of WT pre-processing in converting 1D mechanical signals into a format suitable for 2D CNNs.

Like wavelet transforms, the short time Fourier transform (STFT), another common time-frequency (TF) analysis technique, has been employed for data pre-processing in CNFD. Zhong et al. [123] developed a model combining the STFT to convert vibration signals from the time domain to the time-frequency domain with a CNN to process these transformed signals. Tao et al. [124] proposed an unsupervised fault diagnosis method specifically for rolling bearings, integrating the STFT with a categorical generative adversarial network (CatGAN). The method begins by using the STFT to transform raw 1D vibration signals into 2D time-frequency maps, which are then used as inputs for the CatGAN. These studies demonstrate the effectiveness of STFT pre-processing in converting 1D mechanical signals into a format suitable for 2D CNNs.

While 2D convolutional networks have been widely used for fault classification, a more direct approach involves using 1D convolutional networks to process original time-series data. This section explores applications of 1D convolutional networks for fault classification, focusing on different types of raw sensor data, such as vibration and AE data. Vibration data, due to its accessibility and ease of interpretation, has been the primary source of information for machine fault diagnosis. For instance, Qian et al. [125] developed an adaptive overlapping CNN for diagnosing bearing faults, utilizing raw vibration signals as the training data. Jia et al. [126] introduced a normalized CNN to address the issue of imbalanced fault classification in machinery. They also implemented a neuron activation maximization algorithm to improve insights into the feature learning process of the network. Abdeljaber et al. [127] employed a compact CNN to develop an online condition monitoring approach for detecting faults and assessing the severity of bearings. These studies illustrate the effective use of 1D CNNs in analyzing vibration data, thereby contributing to improvements in fault diagnosis.

Beyond directly using raw vibration data, some researchers have employed features extracted from vibration signals to train 1D CNNs for fault diagnosis. Xie et al. [128] employed a

CNN to extract features from the frequency spectrum of vibration signals. These extracted features were then integrated with energy entropy obtained from EMD and time-domain features to achieve a more comprehensive fault classification. Dong et al. [129] utilized both 1D and 2D CNNs to extract features from the frequency spectrum and the STFT spectrum of vibration signals, respectively. They then combined these extracted features for a more comprehensive approach to monitoring roller bearing degradation.

Beyond vibration data, 1D convolutional networks have also been applied to other mechanical signals for fault classification, including current, AE, and built-in encoder data. Li et al. [130] employed a convolutional network combined with a GRU to thoroughly extract features from AE and vibration data. These features were subsequently combined to diagnose gear pitting faults. Jiao et al. [131] proposed a CNN that utilizes built-in encoder information for intelligent fault diagnosis. They developed a multivariate encoder signal via information fusion to gather extensive mechanical health data.

Unlike fault classification, health prediction focuses on tracking the degradation state of machinery, even in the absence of obvious failures. This area is crucial in machine fault diagnosis, as it empowers maintenance personnel to make early assessments and decisions, preventing potential losses and injury. Ong et al. [132] demonstrated the effectiveness of CNNs for gearbox fault diagnosis, highlighting their superior performance compared to traditional machine learning methods by achieving 96.78% accuracy. Zou et al. [133] tackled the crucial challenge of accurately predicting RUL in PHM by introducing a novel approach called the separable convolution backbone network (SCBNet). Pan et al. [134] proposed a lightweight-CNN (LW-CNN) algorithm for achieving real-time fault diagnosis. This approach minimizes computational requirements, making it suitable for real-time applications. Ma et al. [135] proposed a TimeGAN-based data enhancement method that expands the performance data set, generating synthetic EHA data to compensate for limited real-world data. Balamurgan et al. [136] presented a novel approach for intelligent bearing diagnosis, leveraging a combined CNN-BiLSTM model to improve accuracy in RUL prediction and fault detection. He et al. [137] address the critical challenge of accurately assessing the health status of rotating machinery, integrating multiple tasks and features while ensuring data privacy. They proposed a novel framework that combines dual task feature fusion learning with decentralized swarm learning, offering significant advancements in predictive maintenance. Zhang et al. [138] proposed a novel method that utilizes vibration signals and time series prediction techniques to effectively detect and quantify fault severity. This approach provides a robust and reliable solution for early fault detection. Song et al. [139] presented a novel approach for predicting the RUL of rotating machines, utilizing a shape-constrained transfer temporal transformer network (SC3TN). The study demonstrates the effectiveness of the proposed method in achieving accurate RUL predictions across different machines and operating conditions, offering a significant advancement in predictive maintenance. Lv et al. [140] proposed a framework that offers a comprehensive and innovative solution for RUL predictions of rotating machinery, combining advanced feature extraction, degradation modeling, and adaptive learning techniques. Its ability to accurately predict RUL across different operating conditions and adapt to complex degradation patterns makes it a valuable tool for optimizing maintenance strategies and ensuring the safe operation of rotating machinery systems. Yang et al. [141] proposed a LPST-Net framework that effectively addresses the challenges of complex operating conditions, limited data, and computational overhead, offering a robust and practical approach to enhance the safety and performance of industrial systems. Following this, a CNN is created for adaptive feature learning and condition classification. Table 1 provides a concise summary of reviewed literature.

**Table 1**
Applications of CNNs in fault diagnosis

| S. NO | Machine Components | References |
|---|---|---|
| 1 | Bearings | Wenyi et al. [142], Zhang et al. [143], Zhang et al. [144], Choudhary et al. [145], Zhiyu et al. [146], Yang et al. [147], Amarouayache et al. [148], Han et al. [149], Wang et al. [150], Zhao et al. [151], Wu et al. [152], Zhang et al. [153], Huang et al. [154], Hasan et al. [155], Lu et al. [156] and Shao et al. [157] |
| 2 | Gears | Qui et al. [158], Zhao et al. [159], Liang et al. [160], Cao et al. [161], Li et al. [162], Grezmak et al. [163], Cao et al. [161], Li et al. [164], Chen et al. [165], Li et al. [166], Di et al. [167] and Jamil et al. [168] |
| 3 | Pumps | Tang et al. [169], Tang et al. [170], Yan et al. [171], Wang and Xiang [172], Manikandan and Duraivelu [173], Junior et al. [174], Yang et al. [175], Miao et al. [176], Sunal et al. [177], Anvar and Mohammadi [178], Li et al. [179], Liu et al. [180] and Liu et al. [181] |
| 4 | Motors | Ince et al. [104], Wang et al. [182], Liu et al. [183], Piedad et al. [184], Junior et al. [185], Wang et al. [186], Xu et al. [187], Tran et al. [188], Shao et al. [189], Xiao et al. [190], Kumar et al. [191] and Yang et al. [192] |
| 5 | Engines | Fentaye et al. [193], Zhao et al. [194], Zhang et al. [195], Shahid et al. [196], Wang et al. [197], Zhao and Cai [198], Zhao and Chen [199], Chakrapani and Sugumaran [200], Cui et al. [201], Bai et al. [202] and Li et al. [203] |

This section explored the application of CNNs in machine fault classification and health prediction. CNNs stand out from traditional neural networks due to their appealing characteristics, including sparse local connections, weight sharing, and downsampling, which contribute to efficient feature learning. Furthermore, advanced techniques, such as innovative convolutional architectures, normalization, regularization, and optimization strategies have led to impressive diagnostic results. Moreover, the integration of signal processing methods, such as wavelet transforms and short-time Fourier transforms, enhance the performance of CNNs by pre-processing input data and revealing more fault information. This synergy between the internal advantages of CNNs and external processing aids results in remarkable fault diagnosis capabilities.

A key limitation of current methods is their focus on providing point estimates rather than a distribution of prognostics. This rigid approach lacks flexibility in real-world applications. The inability to quantify prediction uncertainty poses a significant issue, especially for equipment with stringent safety requirements, where understanding the range of potential outcomes is crucial for informed decision-making. Another challenge stems from the difficulty of obtaining sufficient lifetime data in industrial settings. Often, there isn't enough data to train a comprehensive life prediction model. Therefore, researchers should focus on building models using experimental or simulated data and then effectively adapting them to real-world industrial applications. These limitations highlight the need for further research to enhance the robustness and adaptability of CNN-based health prediction models. Addressing uncertainty quantification and data scarcity will be crucial for making these models more reliable and applicable in diverse industrial scenarios.

## 6. Future of CNNs in machinery fault diagnosis

The future of CNNs in machinery fault diagnosis is brimming with exciting possibilities, driven by the relentless pursuit of greater accuracy, robustness, and adaptability. As the research landscape moves forward, there will likely be a convergence of multiple advancements, creating a powerful synergy that propels CNFD to new heights.

### 6.1. Advanced architectures and feature Engineering

The quest for more effective CNN architectures for fault diagnosis will continue. Research will focus on designing deeper and wider networks that capture increasingly complex features and relationships within data itself. This journey will likely involve exploring novel convolutional operations, such as group convolutions, dilated convolutions, and deformable convolutions, each designed to extract specific types of features and enhance the model's understanding of machine health.

Furthermore, the field will see significant progress in feature engineering techniques specifically tailored for fault diagnosis. This will involve the development of domain-specific feature extraction methods that leverage prior knowledge of machine dynamics, physical principles, and fault mechanisms. Such techniques could include:

- **Physically-informed feature extraction:** Integrating physics-based models into the feature extraction process to capture meaningful features that reflect the underlying physical processes of machine operation and fault development. This could involve combining sensor data with physics-based models to generate features that represent key parameters like stress, strain, or vibration modes [204].

- **Multi-modal feature fusion:** Combining features extracted from different sensor modalities, such as vibration, acoustic emission, current, temperature, and pressure, to provide a comprehensive representation of machine health. This would involve developing robust fusion strategies that effectively integrate diverse features and enhance the model's ability to detect complex faults [205].

- **Transfer learning for fault diagnosis:** Leveraging the knowledge learned from similar but different domains or tasks to improve fault diagnosis accuracy for new machines or operating conditions. This could involve transferring pre-trained models or fine-tuning existing models on a smaller dataset specific to the new application.

### 6.2. Hybrid architectures: merging strengths for enhanced performance

The future of CNFD will likely see the emergence of hybrid architectures that combine the strengths of CNNs with other deep learning models, such as RNNs, LSTM networks, and transformer networks. This fusion of architectures will leverage the strengths of each model, resulting in powerful hybrid frameworks that can effectively tackle complex fault diagnosis challenges.

- **CNN-RNN hybrids for time series analysis:** Combining CNNs for feature extraction with RNNs or LSTM networks for capturing temporal dependencies and sequential patterns in time-series data. This could be particularly beneficial for analyzing vibration signals, where identifying changes in the temporal patterns can be crucial for fault detection.

- **Transformer-based fault diagnosis:** Employing transformer networks, known for their ability to capture long-range dependencies and learn contextual relationships in data, for fault diagnosis. Transformers could be used to analyze sensor data from multiple points in time or from different machines to identify subtle patterns that indicate faults [206][207].

### 6.3. Real-time fault diagnosis and prognosis

The goal of real-time fault diagnosis and prognosis will drive significant advancements in CNFD. This will involve developing algorithms and architectures that are computationally efficient and can make rapid predictions.

- **Edge computing for real-time inference:** Deploying CNN models on edge devices, such as sensors and embedded systems, to enable real-time fault detection and condition monitoring. This would eliminate the need for data transmission to cloud servers, enabling faster responses and reducing latency.

- **Model compression and optimization:** Developing techniques for compressing CNN models and optimizing their computational efficiency to enable real-time inference on resource-constrained devices. This could involve using techniques like pruning, quantization, and knowledge distillation to reduce model size and complexity.

- **Adaptive learning for dynamic environments:** Designing CNN models that can adapt to changing operating conditions and environmental variations. This could involve using online learning algorithms or incorporating adaptive mechanisms that adjust model parameters based on new data and changing conditions.

## 6.4. Explainable AI (XAI) for improved transparency and trust:

Explainable AI (XAI) will play a crucial role in building trust and confidence in CNFD models. As CNNs become increasingly complex, understanding their decision-making processes is vital.

- **Feature importance visualization:** Developing techniques to visualize the features that CNNs consider most important for making predictions. This could help engineers understand the model's decision-making process and identify potential biases or weaknesses.

- **Attention mechanisms for interpretability:** Incorporating attention mechanisms into CNN architectures to highlight the parts of the input data that the model focuses on. This could provide insights into the model's reasoning process and facilitate the identification of critical features.

- **Counterfactual explanations:** Generating counterfactual explanations that show how the model's prediction would change if specific features were modified. This could help users understand the model's sensitivity to different inputs and provide valuable insights into its decision-making process.

## 6.5. Integration with advanced control and optimization techniques:

The future of CNFD will involve integrating CNNs with advanced control and optimization techniques to improve machine performance and efficiency. This will enable proactive maintenance, predictive control, and optimized operation.

- **Model predictive control (MPC) with CNFD:** Integrating CNFD models into MPC algorithms to predict future machine states and optimize control actions to minimize wear, improve efficiency, and prevent failures.

- **Adaptive control based on CNFD:** Using CNFD models to continuously monitor machine health and adapt control parameters in real-time to optimize performance and prevent failures.

- **Data-driven optimization of maintenance schedules:** Utilizing CNFD models to predict RUL and optimize maintenance schedules, minimizing downtime and maximizing machine availability.

The future of CNNs in machinery fault diagnosis is filled with exciting prospects. By embracing advanced architectures, hybrid frameworks, real-time capabilities, explainable AI, and integration with control and optimization techniques, CNFD will transform the landscape of machinery health monitoring, leading to safer, more efficient, and more sustainable industrial operations. As researchers navigate the future, it's essential to prioritize ethical considerations, ensuring that CNFD technologies are developed and deployed responsibly, addressing issues of data privacy, model robustness, fairness, and transparency. With continued research and innovation, CNFD will play a pivotal role in the evolution of the industrial landscape, contributing to a future where machines operate with unprecedented reliability, safety, and efficiency.

## 7. Conclusions

This comprehensive examination of CNNs in machine fault diagnosis reveals their significant potential to revolutionize predictive maintenance and enhance industrial reliability. CNNs, with their ability to extract complex features from diverse data sources, have demonstrated remarkable success in fault classification, health prediction, and transfer diagnosis. Their inherent strengths, such as sparse local connections, weight sharing, and downsampling, combined with innovative architectures, normalization, regularization, and optimization techniques, have yielded impressive diagnostic results. However, there are some challenges that remain to be addressed. For instance, the reliance on large datasets with sufficient labeled samples presents a significant hurdle, particularly in real-world scenarios where data collection is often difficult or expensive. Furthermore, the assumption of consistent data distributions between training and testing can limit model performance in industrial settings with fluctuating conditions. Addressing these limitations will require further research in areas such as data augmentation, transfer learning, domain adaptation, and robust model development. Despite these challenges, the future of CNNs in machine fault diagnosis is promising. By continuing to explore advanced architectures, hybrid frameworks, and efficient implementations, researchers can develop more robust and reliable models capable of handling diverse data types and complex operating conditions. Integrating explainable AI (XAI) into these models will further enhance transparency and trust, fostering wider adoption. The integration of CNNs with advanced control and optimization techniques will revolutionize industrial operations, enabling proactive maintenance, predictive control, and the optimization of machine performance. However, ethical considerations, such as data privacy, model robustness, fairness, and transparency, must be addressed to ensure the responsible development and deployment of these powerful technologies. The continued research and innovation of CNN-based fault diagnosis strategies will contribute to a future where machines operate with unparalleled reliability, safety, and efficiency, driving advancements in industries worldwide.

Acknowledgements

"The work is supported by the National Center of Science, Poland, under Sheng2 project No. UMO-2021/40/Q/ST8/00024: NonGauMech - New methods of processing non-stationary signals (identification, segmentation, extraction, modeling) with non-Gaussian characteristics for the purpose of monitoring complex mechanical structures".